\documentclass[journal]{IEEEtran}
\usepackage{amsmath,amsfonts}
\usepackage{algorithmic}
\usepackage{array}
\usepackage{textcomp}
\usepackage{stfloats}
\usepackage{url}
\usepackage{verbatim}
\usepackage{graphicx}
\usepackage{multirow}
\usepackage{tabularx}
\usepackage{amsthm,amsmath,amssymb}
\usepackage{mathrsfs}
\usepackage{hyperref}
\usepackage{xcolor}
\usepackage{threeparttable}
\usepackage{bbm}
\usepackage{caption}
\usepackage{algorithm}
\usepackage{algorithmic}
\usepackage{setspace}
\usepackage{color} 
\usepackage{float}
\usepackage{graphicx}
\usepackage{subfigure}
\usepackage{booktabs}
\usepackage[edges]{forest}
\definecolor{hidden-draw}{RGB}{106,142,189}
\definecolor{hidden-blue}{RGB}{194,232,247}
\definecolor{hidden-orange}{RGB}{217, 232, 252}

\def\BibTeX{{\rm B\kern-.05em{\sc i\kern-.025em b}\kern-.08em
    T\kern-.1667em\lower.7ex\hbox{E}\kern-.125emX}}
\usepackage{balance}
\begin{document}
\title{Parameter-Efficient Fine-Tuning Methods for Pretrained Language Models: A Critical \\Review and Assessment}
\author{Lingling Xu, Haoran Xie, Si-Zhao Joe Qin, Xiaohui Tao, Fu Lee Wang
\thanks{This work was supported by a research grant entitled "Medical Text Feature Representations based on Pre-trained Language Models" (871238) and Faculty Research Grant (DB24A4) at Lingnan University, Hong Kong. \emph{(Corresponding author: Haoran Xie.)}\\
\indent Lingling Xu and Fu Lee Wang are with the Hong Kong Metropolitan University, Hong Kong (email: xxiao199409@gmail.com; pwang@hkmu.edu.hk).\\
\indent Haoran Xie and Si-Zhao Joe Qin are with Lingnan University, Hong Kong (email: hrxie@ln.edu.hk; joeqin@ln.edu.hk). \\
\indent Xiaohui Tao is with University of Southern Queensland, Queensland, Australia (email: xtao@usq.edu.au).\\
%\indent Hong-Ning Dai is with Hong Kong Baptist University, Hong Kong (email: henrydai@hkbu.edu.hk).
}}

%\markboth{IEEE TRANSACTIONS ON PATTERN ANALYSIS AND MACHINE INTELLIGENCE}%
%{How to Use the IEEEtran \LaTeX \ Templates}

\maketitle

\begin{abstract}
With the continuous growth in the number of parameters of transformer-based pretrained language models (PLMs), particularly the emergence of large language models (LLMs) with billions of parameters, many natural language processing (NLP) tasks have demonstrated remarkable success. However, the enormous size and computational demands of these models pose significant challenges for adapting them to specific downstream tasks, especially in environments with limited computational resources. Parameter Efficient Fine-Tuning (PEFT) offers an effective solution by reducing the number of fine-tuning parameters and memory usage while achieving comparable performance to full fine-tuning. The demands for fine-tuning PLMs, especially LLMs, have led to a surge in the development of PEFT methods, as depicted in Fig.~\ref{evolution}. In this paper, we present a comprehensive and systematic review of PEFT methods for PLMs. We summarize these PEFT methods, discuss their applications, and outline future directions. Furthermore, we conduct experiments using several representative PEFT methods to better understand their effectiveness in parameter efficiency and memory efficiency. By offering insights into the latest advancements and practical applications, this survey serves as an invaluable resource for researchers and practitioners seeking to navigate the challenges and opportunities presented by PEFT in the context of PLMs.
\end{abstract}


\section{Introduction}
\IEEEPARstart{T}{ransformer-based} PLMs \cite{devlin-etal-2019-bert,liu2020roberta,radford2019language,raffel2020exploring} have demonstrated remarkable performance across a wide range of NLP tasks. To fully harness the potential of PLMs, fine-tuning is employed to adapt the PLMs to task-specific data to enhance performance on downstream tasks. However, traditional fine-tuning involves updating all the pretrained parameters of PLMs, which is time-consuming and computationally expensive. As the size of PLMs continues to increase, from models like BERT \cite{devlin-etal-2019-bert} with 110 million parameters to T5 \cite{raffel2020exploring} with 770 million parameters, computational resource requirements become a significant challenge. The advent of LLMs \cite{zhang2022democratizing,scao2022bloom,touvron2023llama}, exemplified by Falcon \cite{almazrouei2023falcon} with a staggering 180 billion parameters, further exacerbates the computational demands. To perform task-specific full fine-tuning with Falcon-180B, a minimum of 5120GB of computational resources may be required\footnote{\url{https://huggingface.co/blog/falcon-180b\#hardware-requirements}}. The enormous computational resource requirements are prohibitive for anyone but the superpower players to utilize LLMs for task-specific fine-tuning.

To address this challenge, a prominent method known as PEFT \cite{houlsby2019parameter} has emerged as a viable solution to compensate for the tremendous computational cost of full parameter fine-tuning. PEFT involves employing various deep learning techniques \cite{houlsby2019parameter,li-liang-2021-prefix,hu2022lora} to reduce the number of trainable parameters while still maintaining comparable performance to the full fine-tuning. In addition, PEFT updates only a small number of additional parameters or updates a subset of the pretrained parameters, preserving the knowledge captured by the PLM while adapting it to the target task and reducing the risk of catastrophic forgetting. Furthermore, since the size of the fine-tuned dataset is typically much smaller than the pretrained dataset, performing full fine-tuning to update all the pretrained parameters may lead to overfitting, which is circumvented by the PEFT through selectively or not updating pretrained parameters.

\begin{figure*}[htbp]
  \centering
  \includegraphics[height=0.7\textheight, width=\textwidth]{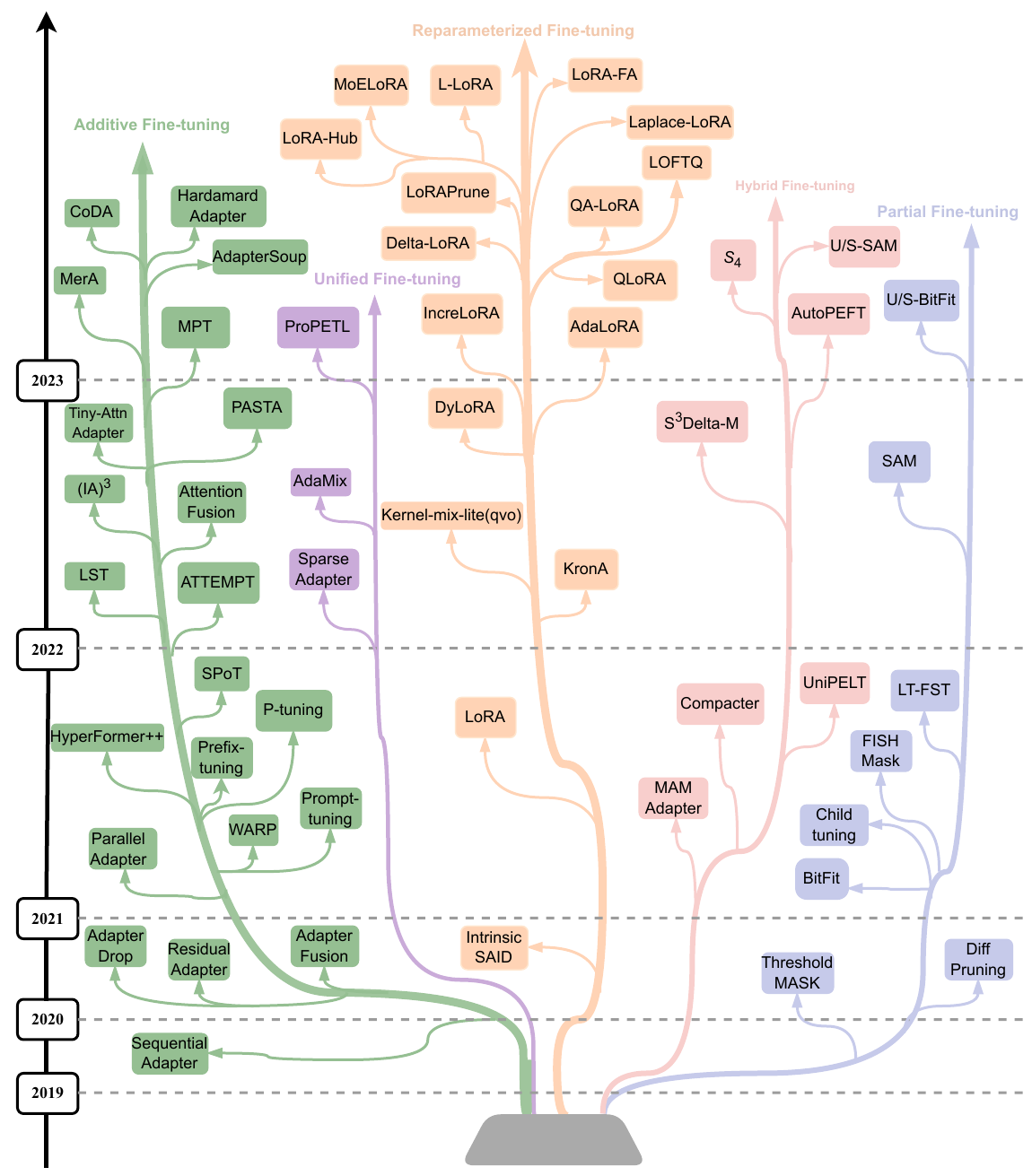}
  \caption{The evolutionary development of PEFT methods in recent years. Models on the same branch have some common features. The vertical position of the models shows the timeline of their release dates. Notably, the year of the paper's initial publication is shown as the reference. For instance, if a paper is published in ACL 2022 but listed on arXiv in 2021, the year 2021 will be considered as the reference date.}
  \label{evolution}
\end{figure*}

Recently, there has been a significant surge in interest regarding PEFT methods, as demonstrated by the growing number of studies depicted in Fig.~\ref{evolution}. This also leads to a few surveys on PEFT approaches for the PLMs. However, the existing surveys have certain limitations. Ding et al. \cite{ding2022delta} conducted a comprehensive study on PEFT methods, but this survey did not cover much of the latest work in the field and only four PEFT methods were quantitatively experimented with. Lialin et al. \cite{lialin2023scaling} delved into the ideas and operational implementations of PEFT methods in detail but do not perform relevant experiments. In this work, we address these gaps comprehensively. We meticulously categorize the PEFT methods, providing detailed explanations of the ideas and specific implementations of each method. We compare the similarities and differences among various types of PEFT methods, facilitating a better understanding of the evolving landscape of PEFT. Moreover, we conduct extensive fine-tuning experiments with 11 representative PEFT methods.

In this paper, we aim to provide a comprehensive and systematic study of PEFT methods for PLMs in NLP. We undertake an in-depth exploration of these PEFT methods and present a comprehensive taxonomy scheme in Section~\ref{section3}. By categorizing PEFT methods into additive fine-tuning, partial fine-tuning, reparameterized fine-tuning, hybrid fine-tuning, and unified fine-tuning, we establish a structured framework for understanding these PEFT approaches, as depicted in Fig.~\ref{PEFT_categorization_of_LLMs}. In Section~\ref{section4}, we conduct quantitative investigations and analyses to assess the performance, parameters efficiency, and memory usage of these PEFT approaches. Our quantitative studies primarily focus on natural language understanding (NLU), machine translation (MT), and natural language generation (NLG) tasks. Additionally, we extensively explore the applications of PEFT in multi-task learning, cross-lingual transfer, and backdoor attack and defense, underscoring its effectiveness. Furthermore, our research also unveils potential directions for future investigations in this rapidly evolving field. To summarize, the main contributions of this survey can be outlined as follows:
\begin{itemize}
\item We present a comprehensive analysis and review of PEFT methods for transformer-based PLMs.
\item We identify the key techniques and approaches employed in PEFT methods, and classify them into additive, partial, reparameterized, hybrid, and unified fine-tuning methods.
\item We conduct extensive experiments to evaluate the effectiveness of several representative PEFT methods, specifically examining their impact on parameter efficiency and memory usage.
\end{itemize} 

\tikzstyle{my-box}=[
 rectangle,
 draw=hidden-draw,
 rounded corners,
 text opacity=1,
 minimum height=1.5em,
 minimum width=5em,
 inner sep=2pt,
 align=center,
 fill opacity=.5,
 ]
 \tikzstyle{leaf}=[my-box, minimum height=1.5em,
 fill=hidden-orange!60, text=black, align=left,font=\scriptsize,
 inner xsep=2pt,
 inner ysep=4pt,
 ]
\begin{figure*}[t]
	\centering
	\resizebox{\textwidth}{!}{
		\begin{forest}
			forked edges,
			for tree={
				grow=east,
				reversed=true,
				anchor=base west,
				parent anchor=east,
				child anchor=west,
				base=left,
				font=\small,
				rectangle,
				draw=hidden-draw,
				rounded corners,
				align=left,
				minimum width=4em,
				edge+={darkgray, line width=1pt},
				s sep=3pt,
				inner xsep=2pt,
				inner ysep=3pt,
				ver/.style={rotate=90, child anchor=north, parent anchor=south, anchor=center},
			},
			where level=1{text width=4.7em,font=\scriptsize}{},
			where level=2{text width=6.8em,font=\scriptsize}{},
			where level=3{text width=6.8em,font=\scriptsize}{},
			[
			PEFT Methods for PLMs, ver
			[
			Additive \\ Fine-tuning 
			[
			Adapter-based \\ Fine-tuning		
			[
			Sequential Adapter~{\cite{houlsby2019parameter},}
			Residual Adapter~{\cite{lin-etal-2020-exploring},}  CoDA~{\cite{lei2023conditional},}
			Parallel Adapter~{\cite{he2022towards},} 
                AdapterDrop~{\cite{ruckle-etal-2021-adapterdrop},} \\
                Tiny-Attn Adapter~{\cite{zhao-etal-2022-tiny},}
                AdapterFusion~{\cite{pfeiffer-etal-2021-adapterfusion},} 
                MerA~{\cite{he2023mera},} 
                Hyperformer++~{\cite{karimi-mahabadi-etal-2021-parameter},}
                AdapterSoup~{\cite{chronopoulou-etal-2023-adaptersoup}}
			, leaf, text width=30em
			]
			]
			[
			Soft Prompt-based \\ Fine-tuning
			[
            WARP~{\cite{hambardzumyan-etal-2021-warp},}
			Promt-tuning~{\cite{lester-etal-2021-power},} Prefix-tuning~{\cite{li-liang-2021-prefix},} P-tuning~{\cite{liu2021gpt},} SPOT~{\cite{vu-etal-2022-spot},} ATTEMPT~{\cite{asai-etal-2022-attempt},} MPT~{\cite{wang2023multitask}}
			, leaf, text width=30em
			]
			]
                [
			Others
			[
                LST~{\cite{sung2022lst},}
			(IA)$^{3}$~{\cite{liu2022fewshot},} PASTA~{\cite{yang-etal-2023-parameter},} 
   AttentionFusion~{\cite{cao-etal-2022-attention},}
                Hadamard Adapter~{\cite{chen2023hadamard}}
			, leaf, text width=30em
			]
			]
			]
			[
			  Partial \\ Fine-tuning
			[
			  Bias Update
			[
			BitFit~{\cite{ben-zaken-etal-2022-bitfit},} 
                U/S-BitFit~{\cite{lawton-etal-2023-neural},} 
                , leaf, text width=30em
			]
			]
                [
			Pretrained Weight \\ Masking
			[
			Threshold-Mask~{\cite{zhao-etal-2020-masking},}
			FISH Mask~{\cite{sung2021training}} 
			, leaf, text width=30em
			]
                ]
                [
			  Delta Weight \\ Masking
			[
			LT-SFT~{\cite{ansell-etal-2022-composable},}
                Child-Tuning~{\cite{xu-etal-2021-raise},}
			Diff Pruning~{\cite{guo-etal-2021-parameter},} 
                SAM~{\cite{fu2023effectiveness}}
			, leaf, text width=30em
			]
			]
			]
			[
			Reparameterized \\ Fine-tuning
			[
			  Low-rank \\ Decomposition
			[
			Intrinsic SAID~{\cite{aghajanyan-etal-2021-intrinsic},}
            LoRA~{\cite{hu2022lora},} KronA~{\cite{edalati2022krona}} 
			, leaf, text width=30em
			]
			]
			[
			  LoRA Derivatives
			[
			Low-rank Adjustment
			[
                DyLoRA~{\cite{valipour-etal-2023-dylora},}
			AdaLoRA~{\cite{zhang2023adaptive},}
			IncreLoRA~{\cite{zhang2023increlora}}
			, leaf, text width=21.6em
			]
			]
                [
			  LoRA-guided Pretrained \\ Weight Update
			[
                Delta-LoRA~{\cite{zi2023delta},}
                LoRAPrune~{\cite{zhang2023pruning}}  
			, leaf, text width=21.6em
			]
			]
                [
			Quantization Adaption
			[
			QLoRA~{\cite{dettmers2023qlora},}
	        QA-LoRA~{\cite{xu2023qa},}
			LOFTQ~{\cite{li2023loftq}} 
			, leaf, text width=21.6em
			]
			]
                [
			  LoRA-based \\ Improvements
			[
			Kernel-mix-lite(qv)/(qvo)~{\cite{chen-etal-2022-empowering},} Laplace-LoRA~{\cite{yang2023bayesian},}
                LoRA-FA~{\cite{zhang2023lora}}          
			, leaf, text width=21.6em
			]
			]
                [
			  LoRA-based \\ Multi-task Fine-tuning
			[
			LoRAHub~{\cite{huang2023lorahub},}
   MoELoRA~{\cite{liu2023moelora},}  
   L-LoRA~{\cite{tang2023parameter}}       
			, leaf, text width=21.6em
			]
			]
			]
			]
			[
			  Hybrid \\ Fine-tuning
			[
			Manual Combination 		
			[
			MAM Adapter~{\cite{he2022towards},} U/S-MAM~{\cite{lawton-etal-2023-neural},} Compacter~{\cite{karimi2021compacter},} UniPELT~{\cite{mao-etal-2022-unipelt},}
			, leaf, text width=30em
			]
			]
   			[
			Automatic Combination		
			[
                AutoPEFT~{\cite{zhou2023autopeft},}
                S$^{3}$Delta-M~{\cite{hu2022sparse},}
			$\mathcal{S}_{4}$~{\cite{chen2023parameterefficient}}
			, leaf, text width=30em
			]
			]
			]
                [
			  Unified \\ Fine-tuning
			[
                AdaMix~{\cite{wang-etal-2022-adamix},}
                SparseAdapter~{\cite{he-etal-2022-sparseadapter},}
			ProPETL~{\cite{zeng-etal-2023-one}}
			, leaf, text width=38.4em
			]
			]
			]
		\end{forest}
  }
\caption{Taxonomy of Parameter-Efficient Fine-Tuning Methods for Pretrained Language Models.}
\label{PEFT_categorization_of_LLMs}
\end{figure*}
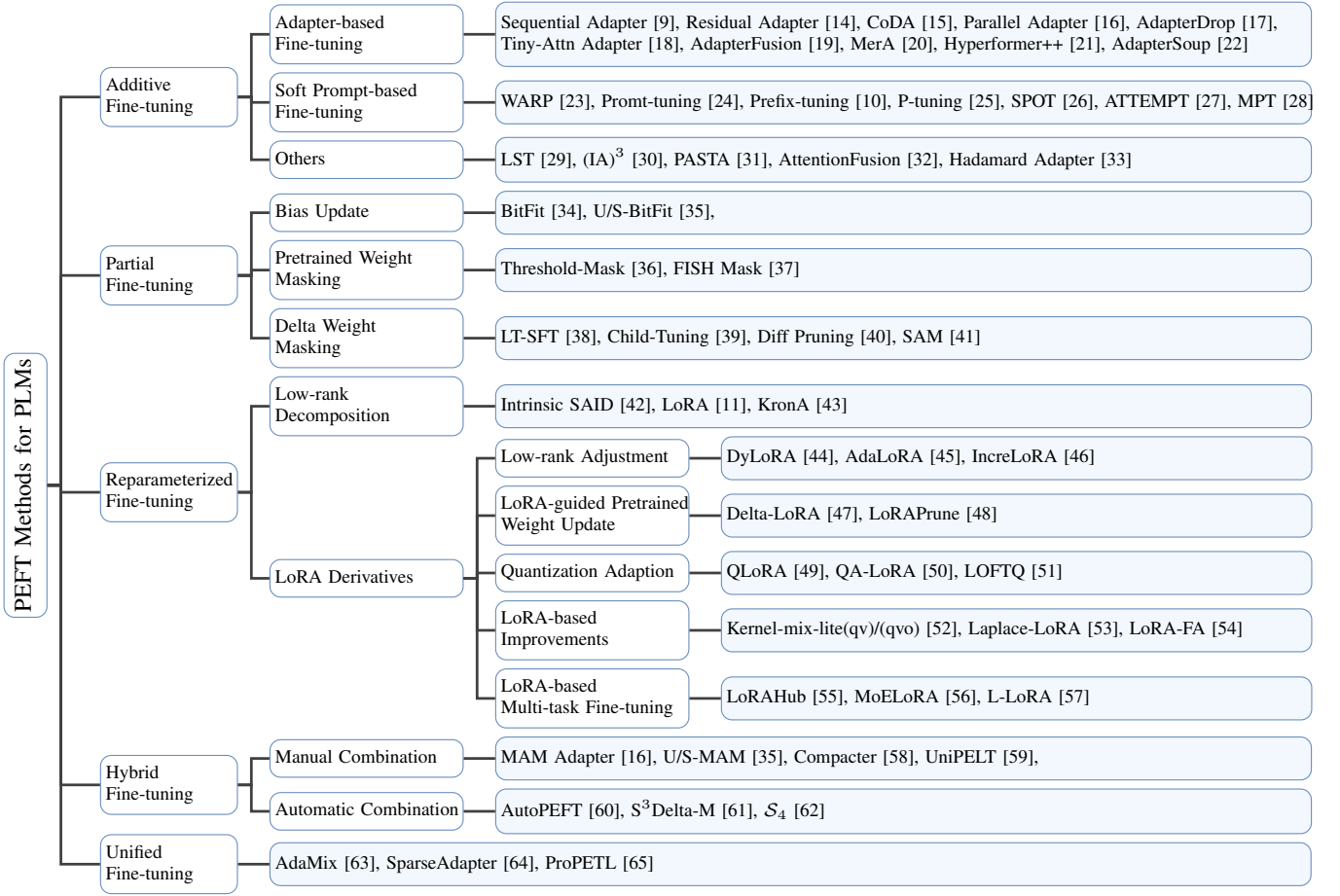

\section{Preliminaries}

\subsection{Transformer}
Transformer \cite{vaswani2017attention} has emerged as a foundational architecture for numerous PLMs, it adopts an encoder-decoder architecture, comprised of a stack of encoder and decoder layers, each equipped with the self-attention mechanism. Both the encoder and decoder in the Transformer architecture consist of a multi-head self-attention layer and a feed-forward network (FFN) layer, interconnected by a residual connection \cite{he2016deep} followed by layer normalization \cite{ba2016layer}. Residual connection allows the model to effectively propagate information from one layer to the subsequent layer without losing valuable information. Layer normalization further stabilizes the training process by normalizing the inputs of each layer.

Multi-head self-attention layer employs the self-attention function with $h$ heads in parallel. For an input sequence $X \in \mathbb{R}^{n \times d}$ with the sentence length $n$ and hidden dimension size of $d$. The query (\textbf{Q}), key (\textbf{K}), and value (\textbf{V}) vectors are the transformation of input sequence $X$,
\begin{equation}
    \textbf{K} = XW_{k} + b_{k}, ~ \textbf{Q} = XW_{q} + b_{q}, ~ \textbf{V} = XW_{v} + b_{v},
\end{equation} where $Q, K, V \in \mathbb{R}^{n \times d}$, $b_{k}$, $b_{q}$ and $b_{v}$ are typically learnable parameter vectors that help model to better capture specific information in the input vector $X$ and adjust the value of the query vector $Q$ to better match the key vector $K$, thereby improving performance of the self-attention mechanism. The self-attention output of input $X$ is computed as:
\begin{equation}
   \text{Attn}(\textbf{Q},\textbf{K},\textbf{V}) = \text{Softmax}(\frac{\textbf{QK}^{T}}{\sqrt{d}})\textbf{V},
\end{equation}
then multi-head self-attention can be described as follows:
\begin{gather}
\text{MHA}(Q, K, V) = \text{Concat}(\text{head}_{1}, \cdots, \text{head}_{h})W^{O},  \\
\text{head}_{i} = \text{Attn}(QW_{Q}^{i}, KW_{K}^{i}, VW_{V}^{i}).
\end{gather}

While the FFN consists of two linear transformations with a non-linear ReLU activation function in between:
\begin{equation}
\text{FFN}(X) = \text{ReLU}(XW_{1} + b_{1})W_{2} + b_{2},
\end{equation}
where $W_{1}$, $b_{1}$, $W_{2}$ and $b_{2}$ are the weight matrices of two linear transformations. Most PEFT methods primarily focus on the self-attention layer and FFN layer, allowing models like encoder-based RoBERTa \cite{liu2020roberta}, encoder-decoder-based T5 \cite{raffel2020exploring}, and decoder-based LLaMA \cite{touvron2023llama} to leverage relevant techniques for parameters reduction.

\subsection{Full Fine-tuning of PLMs}
Full fine-tuning of transformer-based PLMs involves training the entire model, including all layers and parameters, on a specific downstream task using task-specific data. Initially, PLMs are trained on large-scale datasets with unsupervised learning objectives like language modeling or masked language modeling, to learn general language representations \cite{devlin-etal-2019-bert,liu2020roberta,raffel2020exploring,touvron2023llama}. However, these PLMs may not perform optimally when applied to specific tasks like sentiment analysis, question answering, or translation due to a lack of appropriate domain knowledge \cite{xu2023improving,xie2020distant,dabre2019exploiting}. Full fine-tuning provides an effective solution to address this limitation.

During full fine-tuning, the PLM is initialized with pretrained weights and subsequently trained on task-specific data using techniques like backpropagation and gradient descent \cite{hosseini2023towards,amari1993backpropagation}. All model parameters, including pretrained weights, are updated to minimize a task-specific loss that quantifies the disparity between predicted outputs and ground truth. In this way, full fine-tuning enables the model to learn task-specific patterns and nuances from the labeled data, facilitating predictions or outputs tailored to the target tasks \cite{xu2023contrastive}. Notably, full fine-tuning necessitates substantial computational resources and labeled data, as the model is trained from scratch for the specific target task. Moreover, as PLMs grow in size and with the advent of LLMs containing billions of parameters, full fine-tuning places even greater demands on computational resources. In contrast, PEFT methods aim to alleviate these requirements by selectively updating or modifying specific parts of the PLMs while still achieving performance comparable to full fine-tuning \cite{ben-zaken-etal-2022-bitfit,xu-etal-2021-raise}. Furthermore, full fine-tuning may give rise to overfitting when the task-specific dataset is small or when the PLMs are already well-suited to the target task \cite{pfeiffer-etal-2021-adapterfusion,pfeiffer-etal-2020-mad}.

\section{Parameter-Efficient Fine-Tuning Methods}\label{section3}

%We present a detailed introduction to the PEFT methods for PLMs in this section. To facilitate a clearer understanding of PEFT methods, we classify PEFT methods into five main categories: additive, partial, reparameterized, hybrid, and unified fine-tuning, as illustrated in Fig.~\ref{PEFT_categorization_of_LLMs}. 

% This categorization scheme enables a structured and systematic exploration of the various approaches in the PEFT framework, shedding light on their distinctive characteristics and mechanisms.

\subsection{Additive Fine-tuning}
Additive fine-tuning approaches involve introducing new extra trainable parameters for task-specific fine-tuning. We classify additive fine-tuning into three groups: \textbf{Adapter-based Fine-tuning} \cite{houlsby2019parameter,lin-etal-2020-exploring,lei2023conditional,he2022towards,ruckle-etal-2021-adapterdrop,zhao-etal-2022-tiny,pfeiffer-etal-2021-adapterfusion,he2023mera,karimi-mahabadi-etal-2021-parameter,chronopoulou-etal-2023-adaptersoup,zhu2021counter}, in which the adapter module is incorporated into the transformer, allowing for fine-tuning without modifying the pretrained parameters, \textbf{Soft Prompt-based Fine-tuning} \cite{li-liang-2021-prefix,hambardzumyan-etal-2021-warp,lester-etal-2021-power,liu2021gpt,vu-etal-2022-spot,asai-etal-2022-attempt,wang2023multitask}, where soft prompts or prefix vectors are appended to the input embeddings or hidden states during fine-tuning, and \textbf{Others} \cite{sung2022lst,liu2022fewshot,yang-etal-2023-parameter,cao-etal-2022-attention,chen2023hadamard}, in which various methods that introduce supplementary parameters for model fine-tuning are fall into this category.

\subsubsection{Adapters-based Fine-tuning}
\begin{figure*}[t]
\centering
\subfigure[Sequential Adapter]{
\includegraphics[width=2.1in]{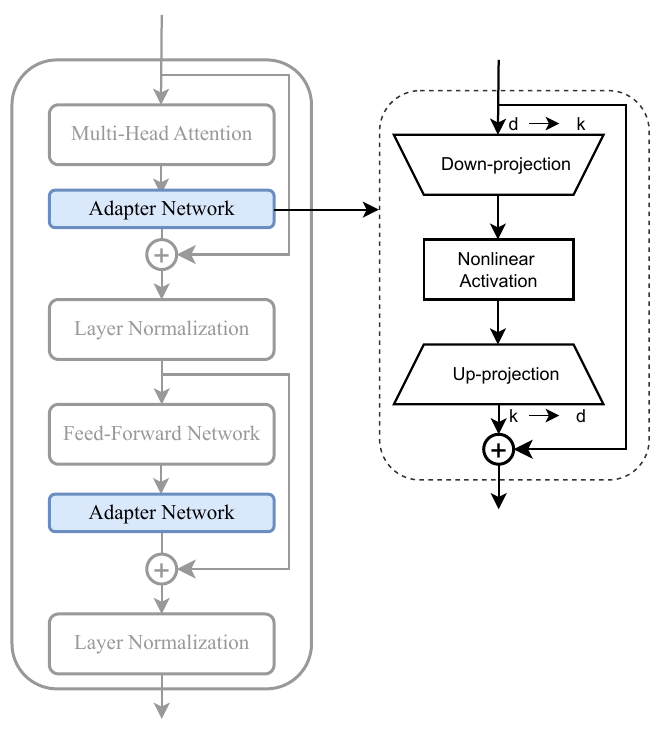}
}
\quad
\subfigure[Prefix-tuning]{
\includegraphics[width=2.1in]{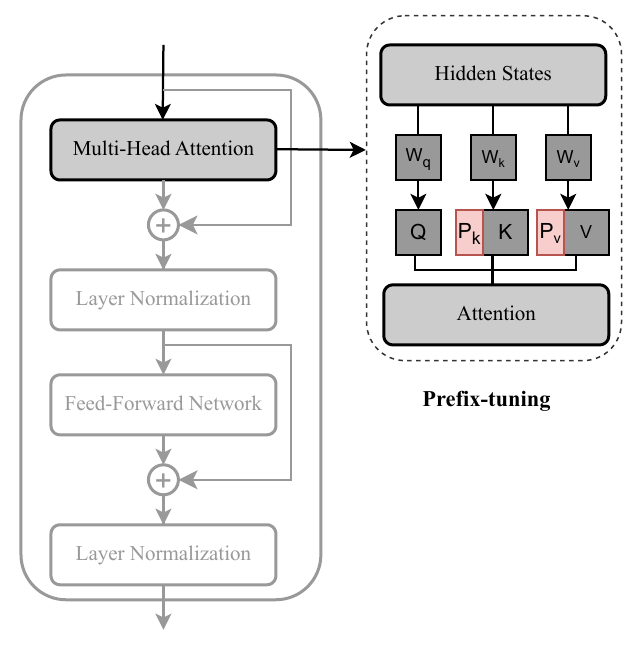}
}
\quad
\centering
\subfigure[LoRA]{
\includegraphics[width=2.1in]{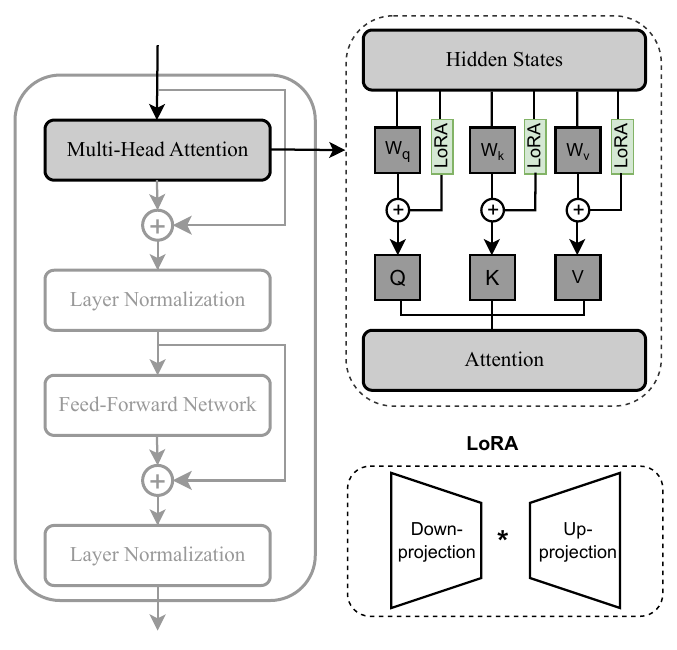}
}
\quad
\caption{The detailed architecture of (a) \textbf{Sequential Adapter}, (b) \textbf{Prefix-tuning}, and (c) \textbf{LoRA}.}
\label{structure}
\end{figure*}

The idea of Adapter is first introduced in multi-domain image classification \cite{rebuffi2017learning}, allowing for the efficient transfer of knowledge across multiple visual domains. \textbf{Sequential Adapter}\cite{houlsby2019parameter} extends and applies it to NLP tasks by inserting the adapter (trainable modules) into the transformer block and fine-tuning the parameters of adapters to make the PLMs adapt to the downstream tasks. Specifically, adapter networks are inserted after the self-attention layer and feed-forward layer of the Transformer sequentially. Each adapter are low-rank module that consists of a down-projection, a non-linear activation function, and an up-projection as well as a residual connection. For the input $X$, the output of a sequential adapter with the ReLU non-linear activation function can be defined with Equation~\ref{adapter}. During fine-tuning, only the parameters of adapter network $W_{up}$ and $W_{down}$ need to be updated to make the PLMs adapt to the specific downstream tasks. The specific architecture of the sequential adapter is presented in Fig.~\ref{structure}. 
\begin{gather}
    X = (\text{ReLU}(X W_{down}))W_{up} + X,
\label{adapter}
\end{gather}
$$W_{down} \in \mathbb{R}^{d \times k}, W_{up} \in \mathbb{R}^{k \times d}.$$

Inspired by sequential adapter, many adapter-based PEFT methods have been proposed. \textbf{Residual Adapter} \cite{lin-etal-2020-exploring} further improves efficiency by inserting the adapter module only after the feed-forward and layer normalization. \textbf{Parallel Adapter} \cite{he2022towards,zhu2021counter} inserts the adapter network in parallel with both the attention layer and the feed-forward layer, allowing for more efficient integration of the adapter module into the transformer. \textbf{AdapterDrop} \cite{ruckle-etal-2021-adapterdrop} removes adapters in each layer of the transformer that are not important to the given task to improve inference efficiency. While \textbf{CoDA} (Condition Adapter) \cite{lei2023conditional} employs parallel adapter for task-specific parameter fine-tuning and remains most pretrained parameters fixed. However, unlike prior methods in which all input tokens are processed with pretrained transformer, CoDA utilizes a router function to select $k$ important input tokens for conditional computation. In this way, CoDA not only enhances parameter efficiency but also inference efficiency. \textbf{Tiny-Attn Adapter} (Tiny-Attention Adapter) \cite{zhao-etal-2022-tiny} introduces a dot-product attention module between the down- and up-projections, which can also be seen as a multi-head attention module with its per-head dimensionality to be extremely small. Moreover, the Tiny-Attn Adapter regards its multiple attention heads as a mixture of experts and averages their weights to further reduce inference costs. Akin to the sequential adapter, the Tiny-Attn Adapter is also injected right after the multi-head attention layer.

\textbf{AdapterFusion} \cite{pfeiffer-etal-2021-adapterfusion} integrates multiple task-specific adapters into a single adapter module, allowing for effective knowledge transfer across related tasks without modifying the original pretrained model. AdapterFusion provides a practical and efficient approach to task composition, enabling the transferability of pretrained models across multiple tasks while minimizing the computational costs associated with fine-tuning the entire model. However, AdapterFusion requires additional trainable parameters in the composition layers, increasing computational costs. \textbf{MerA} (Merging Pretrained Adapters) \cite{he2023mera} adopts summation and averaging strategies to merge the parameters of pretrained adapters without introducing extra trainable parameters. It employs the optimal transport method \cite{solomon2015convolutional,singh2020model} to align the parameters of adapters based on weights and activations, which gives better performance with fewer trainable parameters compared to AdapterFusion. \textbf{Hyperformer++} \cite{karimi-mahabadi-etal-2021-parameter} utilizes the shared hypernetwork \cite{ha2017hypernetworks} to learn task-specific and layer-specific adapter parameters that condition on task and layer id embeddings. By sharing knowledge across tasks via hypernetworks while enabling the model to adapt to each task through task-specific adapters, significantly reducing the number of trainable parameters. \textbf{AdapterSoup} \cite{chronopoulou-etal-2023-adaptersoup} is developed to address cross-domain task adaptation, which first trains multiple adapters based on various domains and then employs domain clustering \cite{aharoni-goldberg-2020-unsupervised} to select the most appropriate top-$k$ adapters for the new domain. Fine-tuning parameters for the new domain in AdapterSoup are determined by calculating the weighted average of the selected $k$ adapters. Apart from cross-domain task adaptation, AdapterSoup can also be used to strengthen in-domain results via weight averaging of adapters trained on the same domain but with different hyperparameters. 

\subsubsection{Soft Prompt-based Fine-tuning}
Soft prompt fine-tuning is a class of methods in which trainable continuous vectors, known as soft prompts, are inserted into the input or hidden state of the model. Unlike manually designed hard prompts, soft prompts are generated by searching for prompts in a discrete token space based on task-specific training data. Soft prompts exhibit more flexibility and adaptability during fine-tuning, as these prompts can be optimized and adjusted based on the specific task and training data.

\textbf{WARP} (Word-level Adversarial RePrograming) \cite{hambardzumyan-etal-2021-warp} inserts special prompt tokens $[P_{1}], [P_{2}], \cdots, [P_{l}]$ and [Mask] token before or after the sentences relying on the prompt template. The training objective is to minimize the cross-entropy loss between the output of MLM and the verbalizer tokens $[V_{1}], [V_{2}], \cdots, [V_{c}]$ for classes $\{1,2,\cdots, c\}$. Only the parameters of $[P_{1}], [P_{2}], \cdots, [P_{l}]$ and $[V_{1}], [V_{2}], \cdots, [V_{c}]$ are trainable, resulting in a significant reduction in the number of fine-tuning parameters. \textbf{Prompt-tuning} \cite{lester-etal-2021-power} incorporates additional $l$ learnable prompt tokens, $P = [P_{1}], [P_{2}], \cdots, [P_{l}]$, into the model input $X \in \mathbb{R}^{n \times d}$ and then concatenates them to generate the final input $\hat{X}$, the new input can be expressed with Equation~\ref{prompt}. During fine-tuning, only the prompt parameters of $P$ are updated through gradient descent, while pretrained parameters remain frozen. Thus, the parameter cost of prompt-tuning is determined by multiplying the prompt length by the token embedding dimension, and extending the prompt length beyond a single token is critical for achieving good performance. 
\begin{equation}
   \hat{X} = \text{Concat}(P, X) = [P, X] \in \mathbb{R}^{(l+n) \times d}.
   \label{prompt}
\end{equation}

\textbf{Prefix-tuning} \cite{li-liang-2021-prefix} proposes to prepend soft prompts $P = [P_{1}], [P_{2}], \cdots, [P_{l}]$ ($l$ denotes the length of the prefix) to the hidden states of the multi-head attention layer, differing from prompt-tuning that adds soft prompts to the input. To ensure stable training, a FFN is introduced to parameterize the soft prompts, as direct optimization of the soft prompts can lead to instability. Two sets of prefix vectors $\hat{P}_{k}$ and $\hat{P}_{v}$ are concatenated to the original key ($K$) and value ($V$) vectors of the attention layer. The self-attention mechanism with prefix-tuning can be represented by Equation~\ref{prefix}. During training, only $\hat{P}_{k}$, $\hat{P}_{v}$, and the parameters of FFN are optimized, while all other parameters of PLMs remain frozen. The structure of prefix-tuning is illustrated in Fig.~\ref{structure}. After training, the FFN is discarded, and only $P_{k}$ and $P_{v}$ are used for inference. \textbf{P-tuning} \cite{liu2021gpt} also considers inserting the soft prompts $[P_{1}], \cdots, [P_{i}], [P_{i+1}], \cdots, [P_{l}]$ into the model input. Nonetheless, P-tuning differs by concatenating these prompts to form a template and maps it to obtain $\{h_{1},\cdots,h_{i},e(x),h_{i+1},\cdots,h_{l}, e(x)\}$, in which $e$ represents pretrained embedding layer. The training goal is to optimize the continuous prompts $\{h_{1}, \cdots, h_{l}\}$. As the weights of PLMs are fixed and only a few parameters need to be fine-tuned, the template can be effectively learned in few-shot learning scenarios. P-tuning employs a bidirectional long short-term memory network (LSTM) with a ReLU-activated multilayer perceptron (MLP) to initialize the embedding of soft prompts through MLP(LSTM($h_{1}, \cdots, h_{i}$): LSTM($h_{i}, \cdots, h_{l}$)).  
\begin{gather}
\label{prefix}
    \text{head} = \text{Attn}(XW_{q}, [\hat{P}_{k},XW_{k}], [\hat{P}_{v}, XW_{v}]), \\
    \hat{P_{k}} = \text{FFN}(P_{k}), \hat{P_{v}} = \text{FFN}(P_{v}).
\end{gather}

\textbf{SPOT} (Soft Prompt Transfer) \cite{vu-etal-2022-spot} is a multitask prompt method that builds upon the prompt-tuning, in which “prompt pertaining” is introduced between PLMs and prompt-tuning of target tasks. There are two variants of SPOT: \emph{generic} SPOT and \emph{targeted} SPOT. \emph{Generic} SPOT first learns a generic prompt on one or more source tasks and then employs the learned prompt to initialize target prompts for specific target tasks. \emph{Targeted} SPOT learns separate prompts for various source tasks, creating a source prompt library. Subsequently, the optimal source prompt, which exhibits higher similarity to the target task embedding, is retrieved and used to initialize the target prompt for the target task. \textbf{ATTEMPT} (ATTEntional Mixtures of Prompt Tuning) \cite{asai-etal-2022-attempt} begins by pretraining transferable soft prompts (source prompts) on large-scale source tasks that possess valuable knowledge applicable to other tasks. The new target prompt is initialized specifically for a given target task. ATTEMPT employs a shared and lightweight network that is trained simultaneously to learn an attention-weighted combination of source prompts and target prompt. This enables modular multi-task learning, as pretrained soft prompts can be flexibly combined, reused, or removed to leverage knowledge from different tasks. \textbf{MPT} (multitask prompt tuning) \cite{wang2023multitask} utilizes multitask data to decompose and distill knowledge from the source prompts to learn a single shared prompt. MPT then learns a multiplicative low-rank matrix to update the shared prompt, efficiently adapting it to each downstream target task. Specifically, MPT assumes that the soft prompts for the $k$-th source task, denoted as $\hat{P}_{k}$, can be decomposed into the shared prompts across all source tasks $P^{*}$ and a task-specific low-rank matrix $W_{k}$. The decomposition is given by $\hat{P}_{k} = P^{*} \odot W_{k} = P^{*} \odot (u_{k} \otimes v_{k}^{T})$, where $\odot$ denotes the Hadamard product, $\otimes$ denotes the Kronecker product, and $u_{k}$ and $v_{k}$ are task-specific vectors for the task $k$. 

\subsubsection{Others}
Apart from adapters family and soft prompts fine-tuning methods, there are some other approaches that also incorporate extra trainable parameters during fine-tuning. They involve adding a ladder side network operating alongside the transformer, introducing an additional diff vector to rescale the attention, incorporating an extra vector to the special token representations, using the late fusion technique to integrate additional attention weight, or combing an extra joint importance weight for each token representation.

\textbf{LST} (Ladder Side-Tuning) \cite{sung2022lst} trains a ladder side network in conjunction with the pretrained network and takes intermediate activations as input via shortcut connections, known as ladders, from pretrained network. Since all training parameters are stored in the ladder side network, back-propagation is achieved through side networks and ladder connections rather than pretrained networks, reducing the number of fine-tuned parameters. In addition, LST further boosts parameter efficiency by utilizing structural pruning \cite{li2017pruning} to retrieve a smaller pruned network to initialize the side network and dropping certain layers of the side network. \textbf{(IA)$^{3}$} (Infused Adapter by Inhibiting and Amplifying Inner Activations) \cite{liu2022fewshot} leverages learned vectors to scale activations, leading to improved performance while introducing a relatively small number of new parameters. (IA)$^{3}$ introduces three learned vectors, $l_{k}$, $l_{v}$, and $l_{ff}$, to rescale the key (K) and value (V) vectors in the attention networks and hidden activations in the position-wise FFN. As a result, the attention output and hidden activation output can be rescaled using the following expressions:
\begin{gather}  
\text{Attn}(Q, K, V) = (\frac{Q(l_{k}\odot K^{T})}{\sqrt{d_{k}}})(l_{v} \odot V), \\
\text{FFN}(X) = (l_{ff} \odot \gamma(XW_{1}))W_{2},    
\end{gather} 
in which $\odot$ represents element-wise multiplication, $W_{1}$ and $W_{2}$ are the weight matrices of FFN, and $\gamma$ is activation function. (IA)$^{3}$ only optimizes three learned vectors $l_{k}$, $l_{v}$, and $l_{ff}$ for each transformer block, resulting in great parameter efficiency. Notably, (IA)$^3$ incurs minimal overhead because $l_k$ and $l_v$ can be seamlessly integrated into the corresponding linear layers, with the only additional overhead arising from $l_{ff}$. \textbf{PASTA} (PArameter-efficient tuning with Special Token Adaptation) \cite{yang-etal-2023-parameter} improves parameter efficiency by modifying the special token representations (e.g., [SEP] and [CLS] in BERT) with an extra special trainable vector before the self-attention layer at each transformer layer. Assuming that the inputs to the transformer layer are denoted as $H = \{h_{i}\}_{i=1}^{N}$, PASTA modifies the inputs as follows:
\begin{equation}  
H_{mod}  = \{h_{i} + m_{i} \}_{i=1}^{N}, 
\end{equation} 
$$m_{i} = e(v_{p}), ~i ~\text{is the}~ p\text{-th special token}; \text{otherwise,}~ m_{i} = 0,$$
$m_{i}$ is the special token adaptation. PASTA enables a remarkable reduction in trainable parameters by training only the trainable vector $e(v_{p})$ to update the representations of special tokens. The reasons for using [CLS] and [SEP] as special tokens in PASTA are that the [CLS] representation provides a global representation of the input text and that the attention scores in PLMs are primarily allocated to the [CLS] or [SEP] tokens across attention heads \cite{clark-etal-2019-bert,kovaleva-etal-2019-revealing}. \textbf{AttentionFusion} \cite{cao-etal-2022-attention} introduces the late fusion technique, which involves combining features or representations from diverse tasks or layers to generate a final joint representation, to adjust the the importance of each token representation. For a given task $t$, let the attention query vector be denoted by $Q^{t}$ and the representation of token $i$ at layer $j$ be $V_{i}^{j}$, then the representation of token $i$ for task $t$, $\hat{V}_{i}^{j}$, is expressed as:
\begin{equation}
\hat{V}_{i}^{j} = \sum_{j}\alpha_{i}^{j}(t)V_{i}^{j},~~~~ \alpha_{i}^{j}(t) = \frac{\text{exp}(Q^{t}V_{i}^{j})}{\sum_{k}\text{exp}(Q^{t}V_{i}^{j})},
\end{equation}
where $\alpha_{i}^{j}(t)$ represents the attention weight of token $i$ at layer $j$ for task $t$. The number of extra parameters that need to be updated in AttentionFusion is determined by the size of the query vector $Q^{t}$, which is the same as the hidden dimension of the pretrained encoder. By employing the attention weight as extra trainable parameters, AttentionFusion adjusts the importance of each token representation dynamically. \textbf{Hadamard Adapter} \cite{chen2023hadamard} is an additive fine-tuning method that introduces a weight vector and a bias vector in each transformer with the same dimensions as the output of the multi-head attention module for fine-tuning. A weight vector and a bias vector are injected right after the multi-head attention layer to perform element-wise multiplication (Hadamard product) with the multi-head attention outputs. Notably, the number of Hadamard adapter is the same as that of the transformer layers in the PLM. During fine-tuning, only the parameters in the Hadamard adapter, layer normalization, and classifier are updated.

\subsection{Partial Fine-tuning}
Partial fine-tuning methods aim to reduce the number of fine-tuned parameters by selecting a subset of pre-trained parameters that are critical to downstream tasks while discarding unimportant ones. We categorize partial fine-tuning methods into three groups: \textbf{Bias Update} \cite{ben-zaken-etal-2022-bitfit,lawton-etal-2023-neural}, in which only the bias term in the attention layer, feed-forward layer and layer normalization of the transformer is updated, \textbf{Pretrained Weight Masking} \cite{zhao-etal-2020-masking,sung2021training}, where the pretrained weights are masked using various pruning criterion, and \textbf{Delta Weight Masking} \cite{ansell-etal-2022-composable,xu-etal-2021-raise,guo-etal-2021-parameter,fu2023effectiveness}, in which delta weights are masked via pruning techniques and optimization approximation. A detailed analysis of pretrained weight and delta weight masking is provided in Table~\ref{partial}.

\begin{table*}
    \caption{The weight update in pretrained weight masking and delta weight masking. $\odot$ denotes the Hadamard product.}
    \centering
    \label{partial}
    \renewcommand{\arraystretch}{1.0}
    \begin{tabular}{|c|c|c|c|}
    \hline
    \textbf{Method} & \textbf{Weight Update} & \textbf{Mask Criterion} & \textbf{Mask Matrix} \\
    \hline
    Threshold-Mask & $\hat{W} =  M \odot W$ & Threshold &  $M = \mathbb{I}_{s_{i,j} > \tau}$ \\
    \hline
    FISH Mask  & $\hat{W} = M \odot W$ & Fisher information &  $M = \mathbb{I}_{top-k({f_{i,j})}}$ \\
    \hline
    LT-SFT &  $\hat{W} = W + M \odot \nabla_{W}\mathcal{L}(W)$  & Absolute difference of parameters &  $M = \mathbb{I}_{top-k(|W_{1}-W_{0}|)}$ \\
    \hline
    Child-Tuning$_{F}$ & $\hat{W} = W - M \odot \eta \nabla_{W} \mathcal{L}(W)$          & Bernoulli distribution &  $M = \{0, 1\}^{n}$\\
    \hline
    Child-Tuning$_{D}$ & $\hat{W} = W - M \odot \eta \nabla_{W} \mathcal{L}(W)$ & Fisher information &  $M = \mathbb{I}_{top-k(f_{i,j})} $      \\
    \hline
    Diff Pruning &  $\hat{W} = W + M \odot \Delta W$   &  Fixed sparsity  & $M = \{0,1\}^{n}$ \\
    \hline
    SAM & $\hat{W} = W + M\Delta W$ & Analytical solution &  $M_{i,j}=0, \forall{i \neq j}; M_{i,i} \in \{0, 1\}$  \\
    \hline
    \end{tabular}
\end{table*}
 
\subsubsection{Bias Update}
\textbf{Bit-Fit} (Bias-term Fine-tuning) \cite{ben-zaken-etal-2022-bitfit} achieves parameter efficiency by only updating the bias terms and the task-specific classification layer while keeping the majority of parameters in the transformer-based PLMs frozen. The bias parameters are involved in the attention layer, where they are involved in calculating query, key, value, and combining multiple attention heads, as well as in the fee-forward and layer normalization layers. Further, \textbf{U/S-BitFit} \cite{lawton-etal-2023-neural} combines NAS algorithm \cite{elsken2019neural} and pruning technique to automatically determine which parameters of the network need to be fine-tuned based on BitFit. \textbf{U-BitFit} (Unstructured BitFit) decides which PEFT parameters to prune based on the first-order approximation of the change in training loss resulting from pruning the PEFT parameter $W$, i.e., $-W \cdot \nabla_{W} \mathcal{L}(W)$. While \textbf{S-BitFit} (Structured BitFit) sums the criterion over the overall bias update $\Delta b$ ($b$ is the bias term). 

\subsubsection{Pretrained Weight Masking}
Pretrained weight masking employs pruning criteria like threshold and Fisher information to measure the importance of pretrained weight to construct a binary mask matrix for weight masking. \textbf{Threshold-Mask} \cite{zhao-etal-2020-masking} utilizes the threshold to construct a binary mask matrix to select pretrained weights $W$ of the attention and FFN layers through element-wise multiplication, expressed as $\hat{W} = W \odot M$ ($\odot$ denotes the Hadamard product). To begin, a random uniformly distributed real-valued matrix $S$, which shares the same dimensions as matrices $W$ and $M$, is created. Subsequently, if an element in $S$ surpasses a predetermined global threshold $\tau$, the corresponding position in the binary mask matrix is assigned a value of 1; otherwise, it is assigned 0. \textbf{FISH Mask} (Fisher-Induced Sparse uncHanging) \cite{sung2021training} uses the Fisher information of pretrained weight to measure their importance and construct a sparse binary mask. FISH Mask selects the top-$k$ parameters with the largest Fisher information to construct the sparse binary mask, where the positions corresponding to the top-$k$ parameters are set to be 1 and the rest are set to 0. Note that $k$ is preset based on the desired mask sparsity level of the mask, and the resulting sparse binary mask can be reused across many subsequent iterations. 

\subsubsection{Delta Weight Masking}
Delta weight masking also employs various pruning techniques and criteria to construct a binary mask matrix to reduce trainable parameters. However, Delta weight pruning typically involves an update at each iteration. \textbf{LT-SFT} (Lottery Ticket Sparse Fine-Tuning) \cite{ansell-etal-2022-composable} is a novel PEFT method inspired by the Lottery Ticket Hypothesis\footnote{Lottery Ticket Hypothesis states that each neural model contains a sub-network (a “winning ticket”) that can match or even outperform the performance of the original model when trained in isolation.} \cite{frankle2018the}. LT-SFT first fine-tunes the PLM on target data using pretrained parameters $W_{0}$ to obtain the fully fine-tuned parameters $W_{1}$, and then identifies the top-$k$ pretrained parameters with the greatest absolute differences ($|W_{1} - W_{0}|$). The top-$k$ parameters are selected for further fine-tuning using binary mask $M$, in which the positions corresponding to the selected $k$ parameters are set to 1 and the remaining positions to 0. LT-SFT then resets the model parameters to their original pretrained weights $W_{0}$ but fine-tunes only the selected $k$ parameters while keeping the remaining parameters frozen, and can be expressed as $\delta = M \odot \Delta W (\Delta W = \nabla_{W}\mathcal{L}(W)$. By iteratively repeating this process, the method gradually fine-tunes only a small fraction of the model's parameters. \textbf{Child-Tuning} \cite{xu-etal-2021-raise} calls the network formed by the parameters to be updated a child network and masks out the gradients of non-child networks to improve parameter efficiency. The parameter updates in Child-Tuning is expressed as $\delta = M \odot \Delta W$ ($\Delta W = \eta \nabla_{W} \mathcal{L}(W)$, $\eta$ denotes learning rate). Child-Tuning provides two variants: Child-Tuning$_{F}$ ($F$ stands for Task-Free) and Child-Tuning$_{D}$ ($D$ stands for Task-Driven). Child-Tuning$_{F}$ generates the binary mask matrix $M$ using Bernoulli distribution with a probability denoted as p$_{F}$. Increasing the value of p$_{F}$ updates a larger number of parameters, and Child-Tuning$_{F}$ is equivalent to full fine-tuning when p$_{F}=1$. In contrast, Child-Tuning$_{D}$ uses Fisher information estimation to identify a subset of parameters (i.e., child network) that are highly correlated with a specific downstream task. The binary mask matrix in Child-Tuning$_{D}$ is constructed by setting the position of the child network to be 1 and the non-child network to be 0. 

\textbf{Diff Pruning} \cite{guo-etal-2021-parameter} introduces a sparse task-specific “diff” vector $\delta$ during fine-tuning while remaining the pretrained model parameters fixed. To make the diff vector $\delta$ sparse, Diff Pruning introduces a learnable binary mask $M$ on the Delta weight and decomposes $\delta = M \odot \Delta W$. The binary mask $M$ is learnable and is used as a regularizer during fine-tuning. It acts as a differentiable approximation to the L0-norm of diff vector $\delta$. This approach is well-suited for multi-task deployment in edge (mobile) applications with limited storage. Significantly, Diff pruning incurs higher memory consumption compared to traditional fine-tuning, which may become problematic as model sizes continue to grow. \textbf{SAM} (Second-order Approximation Method) \cite{fu2023effectiveness} also employs the sparse mask matrix to update the delta weight. However, SAM directly optimizes the approximation function to obtain an analytical solution for the mask matrix, which is then used to update the pretrained weight. Concretely, SAM \cite{fu2023effectiveness} views the PEFT methods as $p$-sparse fine-tuned model by representing fine-tuned parameter as $ W = W_{0} + M \Delta W$, $M$ is a mask matrix, and the optimization problem is 
$$\text{min}_{\Delta W, M}\mathcal{L}(W_{0}+M\Delta W),  ~~~~\ s.t. $$
$$\left\| M \right\|_{0} = \lfloor mp \rfloor, M_{i,j} = 0, \forall{i \neq j}; \text{and} ~~ M_{i,i} \in \{0, 1\}.$$
SAM approximates the loss function using its second-order Taylor expansion as:
\begin{align}
\mathcal{L}(W_{0}+M\Delta W) \approx & \mathcal{L}(W_{0})+\Delta\mathcal{L}(W_{0})^{T}  \Delta \mathcal{L}(W_{0})^{T}M\Delta W \nonumber\\ & + \frac{1}{2}(M\Delta W)^{T}HM\Delta W, 
\end{align}
in which $H$ is the Hessian matrix. In practice, SAM first obtains the gradient $\nabla \mathcal{L}(W_{0})_{i}$ for the $i$-th parameter $W_{i}$, then calculates $\left| \nabla \mathcal{L}(W_{0})_{i}^{2}\right|$, and selects the top $\lfloor mp \rfloor$ delta weight for optimization. 

\subsection{Reparameterized Fine-tuning}
Reparameterized fine-tuning methods utilize low-rank transformation to reduce the number of trainable parameters while allowing operating with high-dimensional matrices (e.g., pretrained weights). We categorize reparameterized fine-tuning methods into two groups: \textbf{Low-rank Decomposition} \cite{hu2022lora,aghajanyan-etal-2021-intrinsic,edalati2022krona}, in which various low-rank decomposition techniques are used to reparameterize the updated matrix, and \textbf{LoRA derivatives} \cite{valipour-etal-2023-dylora,zhang2023adaptive,zhang2023increlora,zi2023delta,zhang2023pruning,dettmers2023qlora,xu2023qa,li2023loftq,chen-etal-2022-empowering,yang2023bayesian,zhang2023lora,huang2023lorahub,liu2023moelora,tang2023parameter}, where a series of PEFT methods are developed based on LoRA. Specific details of $\Delta W$ parameters reparameterization of various approaches can be seen in Table~\ref{delta_reparameterization}.

\begin{table*}
\small
  \caption{Delta weight reparameterization of various reparameterized fine-tuning methods.}
  \centering
  \label{delta_reparameterization}
  \renewcommand{\arraystretch}{1.2}
  \begin{tabular}{|c|c|l|}
    \hline
    \textbf{Method} & \textbf{$\Delta W$ Reparameterization} & \textbf{Notes} \\
    \hline
    Intrinsic SAID & $\Delta W = F(W^{r})$ & $ F: \mathbb{R}^{r} \rightarrow \mathbb{R}^{d}$, $W^{r} \in \mathbb{R}^{r}$ is parameters to be optimized, and $r \ll d$. \\
    \hline
    LoRA & $\Delta W = W_{dowm}W_{up}$ & $W_{down} \in \mathbb{R}^{k \times r}$, $W_{up} \in \mathbb{R}^{r \times d}$, and $r \ll \{k, d\}$. \\
    \hline
    KronA & $\Delta W = W_{down} \otimes W_{up}$ & $\text{rank}(W_{down} \otimes W_{up}) = \text{rank}(W_{down}) \times \text{rank}(W_{up})$.\\
     \hline
    DyLoRA & $\Delta W = W_{down \downarrow b}W_{up \downarrow b}$ &  $W_{down \downarrow b} = W_{down}[:b,:]$, $W_{up \downarrow b} = W_{up}[:,:b]$, $b \in \{r_{min}, \cdots, r_{max}\}$. \\
    \hline
    AdaLoRA & $\Delta W = P\Lambda Q$ & $PP^{T}=P^{T}P=I=QQ^{T}=Q^{T}Q$, $\Lambda = \operatorname{diag}(\sigma_1, \sigma_2, \ldots, \sigma_r)$. \\
    \hline
    IncreLoRA & $\Delta W = W_{down} \Lambda W_{up}$ & $\Lambda = [\lambda_1,\lambda_2,\cdots,\lambda_r]$ with $\lambda_{i}$ could be an arbitrary constant. \\
    \hline
    DeltaLoRA & $\Delta W = W_{down}W_{up}$  &  $W^{(t+1)} \leftarrow W^{(t)} + (W_{down}^{(t+1)}W_{up}^{(t+1)} - W_{down}^{(t)}W_{up}^{(t)})$.\\
    \hline
    LoRAPrune &  $\Delta W = W_{down}W_{up} \odot M$ & $\delta = (W + W_{down}W_{up}) \odot M$, $M \in \{0, 1\}^{1 \times G}$, $G$ is group number. \\
    \hline
    QLoRA &  $\Delta W = W_{down}^{BF16}W_{up}^{BF16}$ & \scriptsize{$Y^{BF16} = X^{BF16}\text{doubleDequant}(c_{1}^{FP32},c_{2}^{FP8}, W^{NF4}) + X^{BF16}W_{down}^{BF16}W_{down}^{BF16}$.} \\
    \hline
    QA-LoRA & $\Delta W = W_{down}W_{up}$ & $W_{down} \in \mathbb{R}^{k \times r}$, $W_{up} \in \mathbb{R}^{r \times L}$, $L$ is the quantization group number of $W$.\\
    \hline
    LOFTQ &  $\Delta W = \text{SVD}(W-Q_{t})$  & $Q_{t} = q_{N}(W-W_{down}^{t-1}W_{up}^{t-1})$, $q_{N}$ is $N$-bit quantization function.\\
    \hline
    Kernel-mix &  \scriptsize{$\Delta W^{h}= (B_{LoRA}, B^{h}) \left(\begin{array}{c}A_{LoRA}^{h}\\A^{h}  
    \end{array}\right)$}  & $B_{LoRA}$ is shared across all heads, $B^{h}$, $A^{h}$ provide rank-$r$ update in each head. \\
    \hline
    LoRA-FA  & $\Delta W = W_{down}W_{up} = QRW_{up}$ & $W_{down}$ is frozen, and only update $W_{up}$.\\
    \hline
  \end{tabular}
\end{table*}

\subsubsection{Low-rank Decomposition}
This involves finding a lower-rank matrix that captures the essential information of the original matrix while reducing computational complexity and memory usage by reparameterizing the updated delta weight. Reparameterization covers transforming the delta weight matrix into a low-rank representation using methods such as Fastfood transformation, low-rank down-up projection, or Kronecker product projection.

\textbf{Intrinsic SAID} (Structure-Aware Intrinsic Dimension) \cite{aghajanyan-etal-2021-intrinsic} leverages the concept of intrinsic dimensionality to reduce the number of parameters during fine-tuning. The intrinsic dimensionality refers to the minimum dimensionality required to solve a high-dimensional optimization problem. In the context of PLMs, measuring the intrinsic dimensionality helps estimate the minimum number of parameters needed to adapt to new tasks. Instead of optimizing the empirical loss in the original parameterization, Intrinsic SAID fine-tunes the model by reparametrization the model in a lower-dimensional space, i.e., $\Delta W = F(W^{r})$, in which $W^{r}$ is the parameter to be optimized and $F: \mathbb{R}^{r} \rightarrow \mathbb{R}^{d}$ is a Fastfood transform\footnote{Fastfood transform is a computationally efficient dimensionality expansion method.} \cite{le2013fastfood} that projects parameters from low-dimensional $r$ to high-dimensional $d$. However, Intrinsic SAID is not practical for fine-tuning larger networks due to the $\mathcal{O}(d)$ memory complexity of the Fastfood transform and the need to update all of the model's parameters. 

Inspired by Intrinsic SAID, \textbf{LoRA} (Low-Rank Adaptation) \cite{hu2022lora} introduces two trainable low-rank matrices for weight update. In LoRA, a down-projection matrix and an up-projection matrix are utilized in parallel with the query (Q), key (K), and value (V) matrices in the attention layer of the transformer, shown in Fig.~\ref{structure}. For a pretrained weight matrix $W \in \mathbb{R}^{d \times k}$, LoRA updates $W$ using low-rank decomposition $\Delta W = W_{down}W_{up}$. During training, the weights of PLM are frozen, and only the low-rank matrices of LoRA, i.e., $W_{down}\in \mathbb{R}^{d \times r}$ and $W_{up} \in \mathbb{R}^{r \times k}$ are fine-tuned ($r \ll \{d, k\}$). During inference, the LoRA weights are merged with the original weight matrix of the PLMs without increasing the inference time. Practically, a scaling factor ($s = 1/r$) is added to the LoRA module. \textbf{KronA} (Kronecker Adapter) \cite{edalati2022krona} is structurally similar to LoRA but replaces the low-rank decomposition in LoRA with Kronecker product decomposition, $\Delta W = W_{down} \otimes W_{up}$. Kronecker product decomposition maintains the rank of the input matrix (i.e., $\text{rank}(A \otimes B) = \text{rank}(A) \times \text{rank}(B)$), ensuring that important information is preserved during the adaptation process. Moreover, Kronecker product can speed up computation and reduce the number of required floating-point operations (FLOPS) by avoiding the explicit reconstruction of the Kronecker product matrix. KronA has two variants: KronA$_{B}$ and KronA$_{\text{res}}^{B}$. KronA$_{B}$ inserts the KronA module in parallel to the FFN layer, while KronA$_{\text{res}}^{B}$ inserts the KronA module alongside the FFN layer and incorporates a learnable residual connection. 

\subsubsection{LoRA Derivatives}
LoRA derivatives refer to a series of PEFT methods that are improved based on LoRA, including \textbf{Low-Rank Adjustment} \cite{valipour-etal-2023-dylora,zhang2023adaptive,zhang2023increlora}, where different methods are developed to adjust the rank of LoRA dynamically, \textbf{LoRA-guided Pretrained Weight Update} \cite{zi2023delta,zhang2023pruning}, in which LoRA is used to guide the update of pretrained weight, \textbf{Quantization Adaption} \cite{dettmers2023qlora,xu2023qa,li2023loftq}, in which various quantization techniques are proposed to improve the high precision fine-tuning and inference of LoRA, \textbf{LoRA-based Improvements} \cite{chen-etal-2022-empowering,yang2023bayesian,zhang2023lora}, in which several novel technique are incorporated into LoRA for improvements, and \textbf{LoRA-based Multi-task Fine-tuning} \cite{huang2023lorahub,liu2023moelora,tang2023parameter}, where multiple LoRA modules are combined for cross-task transfer to fine-tune model on a novel task.

\textbf{Low-rank Adjustment.} \textbf{DyLoRA} (Dynamic LoRA) \cite{valipour-etal-2023-dylora} is introduced to overcome two limitations of LoRA: (a) LoRA's rank is fixed and prevents any changes after training (b) determining the optimal rank for LoRA requires exhaustive search and considerable effort. DyLoRA trains LoRA modules for a range of ranks instead of a single rank, allowing for adaptability. DyLoRA addresses these limitations during training by sorting the representations learned at various ranks. Specifically, DyLoRA operates within the range of ranks denoted as $r \in [r_{min}, r_{max}]$ for a series of iterations. In each iteration, DyLoRA randomly selects a specific rank $b$ from $\{r_{min}, \cdots, r_{max}\}$. It then truncates the down-projection matrix as $W_{down \downarrow b} = W_{down}[:b,:]$ and the up-projection matrix as $W_{up \downarrow b} = W_{up}[:,:b]$ and only update truncated parameter matrices $W_{down \downarrow b}$ and $W_{up \downarrow b}$. The parameter updates in each iteration of DyLoRA could be expressed as $\Delta W = W_{down \downarrow b}W_{up \downarrow b}$. By allowing dynamic low-rank adaptation and search-free low-rank adaptation, DyLoRA reduces the computational cost and training time required to identify the optimal rank for a particular task. \textbf{AdaLoRA} (Adaptive Low-Rank Adaptation) \cite{zhang2023adaptive} extends LoRA by dynamically adjusting the rank of matrices to control the allocation budget. In AdaLoRA, the incremental update $\Delta W$ is reparameterized using singular value decomposition (SVD) and then truncates the smallest singular values, i.e., $\Delta W = P\Lambda Q$. Both $P$ and $Q$ are orthogonal matrices, and $\Lambda$ is a diagonal matrix containing the singular values $\{\sigma_1, \sigma_2, \ldots, \sigma_r\}$. Here, $r$ represents the rank of the matrix $\Lambda$. During training, $P$ and $Q$ are initialized with Gaussian distribution with a regularizer to ensure the orthogonality, while $\Lambda$ is initialized with zero and iteratively pruned to adjust the rank. AdaLoRA employs the sensitivity-base importance scoring \cite{molchanov2019importance,sanh2020movement} with a new metric to prune the singular values of unimportant updates to update the $\Lambda$. By doing this, AdaLoRA effectively improves parameter efficiency and allocation budgets. \textbf{IncreLoRA} \cite{zhang2023increlora} dynamically incorporates trainable parameters into LoRA by increasing their ranks, guided by importance scores assigned to each module during training. The allocation process assigns lower ranks, possibly 0 to indicate no parameter updates, to less important modules, while allocating higher ranks to more important modules. The parameter updates in IncreLoRA can be expressed as $\Delta W = W_{down} \Lambda W_{up}$, in which $\Lambda = [\lambda_1,\lambda_2,\cdots,\lambda_r]$ is a diagonal matrix with $\lambda_{i}$ could be any arbitrary constant, $r$ is the rank of the each LoRA module. Besides, an upper bound on the rank is set for each module to control the parameter growth. Additionally, IncreLoRA introduces a unique pretraining technique called “advance learning”, which ensures that the newly added parameters in each module begin with favorable initial states. In this way, it prevents insufficient training of subsequently added parameters, allowing for effective utilization of the incremental parameter allocation. Unlike LoRA, which operates on the query (Q), key (K), and value (V) projection modules of the attention layer, the parameter updates are applied to all linear layers in IncreLoRA. 

\textbf{LoRA-guided Pretrained Weight Update.} \textbf{Delta-LoRA} \cite{zi2023delta} updates the pretrained weight $W$ as well as two low-rank matrices $W_{down}$ and $W_{up}$, while using the same memory as the original LoRA. The two low-rank matrices $W_{down}$ and $W_{up}$ are automatically updated as usual. The pretrained weight, however, leverages the mathematical property that $\nabla_{W} \mathcal{L}(W, W_{down}, W_{up}) = \nabla_{W_{down}W_{up}} \mathcal{L}(W, W_{down}, W_{up})$ (it is achieved by removing the dropout layer in the original LoRA module) for parameters update. Specifically, $W$ is updated with the delta of the product of two low-rank matrices in consecutive iterations, i.e., $W \leftarrow W + \Delta W_{down}W_{up} = W + (W_{down}(t+1)W_{up}(t+1) - W_{down}(t)W_{up}(t))$. \textbf{LoRAPrune} \cite{zhang2023pruning} introduces a LoRA-guided pruning criterion, which utilizes the weights and gradients of LoRA instead of the gradients of pretrained weights for importance estimation to prune parameters of LoRA and pretrained weights. To address the substantial memory overhead associated with unstructured pruning and dependency-aware structured pruning, LoRAPrune devises a structured iterative pruning procedure that selectively eliminates redundant channels and heads. LoRA-guided pruning criterion involves using low-rank matrices $W_{down}$ and $W_{up}$, along with their corresponding gradients $\nabla_{W_{down}}$ and $\nabla_{W_{up}}$, to calculate the importance score\footnote{Importance score $I$ is calculate via $I = \nabla_{W} \odot W, \nabla_{W} \approx W_{down} \cdot \nabla_{W_{up}} + \nabla_{W_{down}} \cdot W_{up} - \nabla_{W_{down}} \cdot \nabla_{W_{up}}$.}. This score determines which weights are deemed unimportant and subsequently removed. Notably, LoRAPrune not only prunes structured weights, such as heads and channels, from the pretrained weights, but also prunes the corresponding weights in the LoRA, i.e., $\delta = (W + W_{down}W_{up}) \odot M$, $M \in \{0, 1\}^{1 \times G}$, $G$ is the group number. Binary mask $M$ is set to 0 when the corresponding group is unimportant, and 1 when it is important. Therefore, after pruning and fine-tuning, the LoRA weights can seamlessly merge with the pretrained weights, ensuring that no additional computations are necessary during inference.

\textbf{Quantization Adaption.} \textbf{QLoRA} \cite{dettmers2023qlora}, a quantized variant of LoRA, effectively addresses the limited computational resource of LoRA for fine-tuning LLMs by quantizing the transformer model to 4-bit NormalFloat (NF4) precision with double quantization processing, and using a paged optimizer to deal with memory spikes. NF4 is a new data type that is theoretically optimal for normally distributed weights. Although QLoRA quantizes pretrained weight $W$ from FP16 into NF4 so that LLMs can be fine-tuned with fewer GPUs, the auxiliary weight of LoRA matrix $W_{down}W_{up}$ makes the final weight return to FP16 again after fine-tuning. To this end, \textbf{QA-LoRA} (Quantization-Aware Low-rank Adaption) \cite{xu2023qa} employs group-wise quantization with low-rank adaptation to the pretrained weight $W$, in which each column of $W$ is partitioned into $L$ groups for quantization. In this way, QA-LoRA ensures that pretrained weights $W$ and auxiliary weights are integrated into a quantized form after fine-tuning, resulting in a faster and more accurate computation during inference. While \textbf{LOFTQ} (LoRA-Fine-Tuning-aware Quantization) \cite{li2023loftq} applies an N-bit quantized weight $Q$ and low-rank approximation $W_{down} \in \mathbb{R}^{d_{1} \times r}$, $W_{up} \in \mathbb{R}^{d_{2} \times r}$ to approximate the original high-precision pretrained weight $W \in
\mathbb{R}^{d_{1} \times d_{2}}$ as the initialization of LoRA fine-tuning. Such an initialization alleviates the quantization discrepancy in QLoRA and significantly improves the generalization in downstream tasks.

\textbf{LoRA-based Improvements.} \textbf{Kernel-wise Adapter} \cite{chen-etal-2022-empowering} treats the different attention heads in the transformer as independent kernel estimators and utilizes the kernel structure in self-attention to guide the assignment of tunable parameters. LoRA is used as the underlying model to combine kernel-wise adaptation for its flexibility in parameter assignment for different weight matrices. Kernel-wise adapter has two variants: Kernel-mix-lite (qv) and Kernel-mix-lite (qvo). Kernel-mix-lite (qv) provides a lightweight solution for scenarios with limited parameter budgets, while Kernel-mix (qvo) is suitable for scenarios with intermediate parameter budgets. The suffix (qv) means that the method will adjust $W_{q}$ and $W_{v}$, while the suffix (qvo) means that the method will modify $W_{q}$, $W_{v}$, and $W_{o}$. \textbf{Laplace-LoRA} \cite{yang2023bayesian} incorporates Bayesian inference into the LoRA parameters to address the issue of overconfidence and improve calibration. A key challenge lies in obtaining the posterior distribution for Bayesian inference, which is resolved by using Laplace approximation \cite{mackay1992practical}. Laplace-LoRA can be viewed as an approximation of the posterior distribution over LoRA parameters using Laplace approximation. Hence, Laplace-LoRA maintains existing pretraining and fine-tuning procedures while reducing the dimensionality of Bayesian inference. \textbf{LoRA-FA} (LoRA with Frozen-A) \cite{zhang2023lora} is proposed to reduce the expensive activation memory of LoRA without introducing any computational overhead. LoRA-FA keeps the pretrained weight $W$ and down-projection matrix $W_{down}$ frozen and only updates the up-projection matrix $W_{up}$. $W_{down}$ is decomposed into $Q$ and $R$ via QR decomposition, and $\Delta W = W_{down}W_{up} = QRW_{up} = Q\hat{W}_{up} = \sum_{i=1}^{r}Q_{:,i}\hat{W}_{up,i,:}$, in which $\{Q_{:,i}\}_{i=1}^{r}$ are orthogonal unit vectors ($r$ is the rank of $W_{down}$). Thus, $\Delta W$ is a combination of $r$ orthogonal vectors, limiting the change of weight residing in a low-rank space. Consequently, there is no need to store full-rank input activations simultaneously, alleviating the memory burden associated with activation storage.

\textbf{LoRA-based Multi-task Fine-tuning.} \textbf{LoRAHub} \cite{huang2023lorahub} leverages a composition of multiple trained LoRA modules for cross-task transfer to fine-tune the model on new tasks. Specifically, LoRAHub trains task-specific LoRA modules in a variety of tasks to obtain a synthesized module, $\hat{m} = (w_{1}W_{down}^{1} + \cdots + w_{N}W_{down}^{N})(w_{1}W_{up}^{1} + \cdots + w_{N}W_{up}^{N})$, which is then amalgamated with the LLMs to adapt the new task. Thus, the objective of LoRAHub is to find the best weight set $\{w_{1}, w_{2}, \cdots, w_{N}\}$, which is achieved by the gradient-free combinatorial optimization approach Shiwa \cite{liu2020versatile}. \textbf{MOELoRA} \cite{liu2023moelora} combines LoRA with mixture-of-experts (MoE) for multi-task fine-tuning, in which each expert is a LoRA module for learning task-specific knowledge. Additionally, MOELoRA devises a task-motivated gate function to produce distinct fine-tuned parameters for various tasks. \textbf{L-LoRA} (Linearized LoRA) \cite{tang2023parameter} is a linearized PEFT method to improve the multi-task fusion capability of fine-tuned task-specific models with low computation costs. L-LoRA constructs a linear function using a fir-order Taylor expansion, as illustrated in Equation~\ref{linerization}.
In L-LoRA, only the linearized LoRA modules are fine-tuned in the tangent space, incurring fewer trainable parameters compared to LoRA. For the multi-task fusion methods, simple average, task arithmetic \cite{ilharco2023editing,zhang2023composing}, ties-merging \cite{yadav2023resolving}, and LoRAhub \cite{huang2023lorahub} are employed for multi-task fusion. 
\begin{align}
    f_{\theta_{0}}(x;\phi(t)) \approx & f_{\theta_{0}}^{\text{lin}}(x;\phi(t)) = f_{\theta_{0}}(x;\phi(0)) \nonumber\\ & + \nabla_{\phi} f_{\theta_{0}}(x;\phi(0))^{T}(\phi(t)-\phi(0)).
    \label{linerization}
\end{align}

\subsection{Hybrid Fine-Tuning}
Hybrid fine-tuning approaches aim to combine various PEFT approaches, such as adapter, prefix-tuning, and LoRA, to leverage the strengths of each method and mitigate their weaknesses. By integrating different features of PEFT methods, hybrid fine-tuning achieves improved overall performance compared to individual PEFT methods. These works are classified into two approaches: \textbf{Mannual Combination} \cite{he2022towards,lawton-etal-2023-neural,karimi2021compacter,mao-etal-2022-unipelt}, in which multiple PEFT methods are combined manually by sophisticated design, and \textbf{Automatic Combination} \cite{zhou2023autopeft,hu2022sparse,chen2023parameterefficient}, where various PEFT methods are incorporated automatically via structure search.

\subsubsection{Manual Combination}
Manual combination mainly involves integrating the structure or features of one PEFT method into another PEFT method to enhance performance while achieving parameter efficiency. \textbf{MAM Adapter} (Mix-And-Match Adapter) \cite{he2022towards} is the combination of scaled parallel adapter and prefix-tuning, which employs prefix-tuning with smaller bottleneck dimensions at the attention layer and allocates more parameter budget to modify the representation of FFN using the scaling parallel adapter. Scaled parallel adapter denotes the parallel adapter with a scaling factor to adjust the adapter output. Concretely, the output of the MAM adapter $h$ can be expressed with $h = \text{LN}(X + \text{scale} * \text{FFN}(\text{LN}(\text{Attn}([P_{k}, X]) + [P_{v}, X])))$ for the input $X$. Further, \textbf{U-MAM} (Unstructured MAM) and \textbf{S-MAM} (structured MAM) \cite{lawton-etal-2023-neural} are proposed by combining NAS algorithm \cite{elsken2019neural} and pruning technique to automatically determine which parameters of the network need to be fine-tuned based on MAM adapter. NAS algorithm takes the maximum number of parameters required for PEFT architectures as input and applies the pruning operation to reduce trainable parameters. The criteria for deciding which PEFT parameters to prune are based on the first-order approximation of the change in training loss resulting from pruning the PEFT parameter $W$, i.e., $-W \cdot \nabla_{W} \mathcal{L}(W)$. U-MAM directly employs this criterion to prune the parameters in MAM, while S-MAM sums the criterion over each column of $W_{down}$. 

\textbf{Compacter} \cite{karimi2021compacter} is developed based on adapters, low-rank optimization, and a parameterized hypercomplex multiplication (PHM) layer \cite{zhang2021beyond}. It follows a similar structure to adapters, consisting of a down-projection, a nonlinear activation function, and an up-projection. However, Compacter replaces the down-projection and up-projection in the adapters with the low-rank parameterized hypercomplex multiplication (LPHM) layer, which is an extension of PHM that incorporates low-rank optimization. Structurally, PHM layer resembles a fully connected layer, but with the learned $W$ represented as a sum of Kronecker products, i.e., $W= \sum_{i=1}^{n}A_{i} \otimes B_{i}$. Notably, when the weights of down-projection and up-projection are calculated as in that of the PHM layer, $A_{i}$ is a shared parameter across all adapter layers, while $B_{i}$ represents adapter-specific parameters. This kind of adapter is called \textbf{PHM Adapter}. Similarly, Compacter obtains the weight matrix in each LPHM layer utilizing the sum of Kronecker products, but Compacter reparameterizes $B_{i}$ as the product of two independent ranks with one weight, and the weight matrix in Compacter is calculated as follows:
\begin{equation}
    W= \sum_{i=1}^{n}A_{i} \otimes B_{i} = \sum_{i=1}^{n}A_{i} \otimes (s_{i}t_{i}^\mathrm{T}).
\label{compacter}
\end{equation}
$$ W \in \mathbb{R}^{k \times d}, A_{i} \in \mathbb{R}^{n \times n}, B_{i} \in \mathbb{R}^{\frac{k}{n} \times \frac{d}{n}}; s_{i} \in \mathbb{R}^{\frac{k}{n} \times r}, t_{i} \in \mathbb{R}^{r \times \frac{d}{n}}.$$
Compacter++ is a variant of Compacter that inserts a Compacter layer after the FFN layer of each transformer module and requires fewer parameters to be updated than Compacter. 

\textbf{UniPELT} \cite{mao-etal-2022-unipelt} incorporates sequential adapter, prefix-tuning, and LoRA via a gating mechanism. In UniPELT, adapters are added after the feed-forward layer, prefix-tuning is employed to the key ($K$) and value ($V$) vectors of the multi-head attention layer, and LoRA is used in attention matrices of $W_{q}$ and $W_{v}$ of the transformer. Each PEFT module is equipped with a gating mechanism composed of a linear function with the dimension of the output being 1, a sigmoid function, and a mean function. The gating mechanism controls the activation of each submodule, dynamically assigning higher weights to submodules that make positive contributions to a given task. The trainable parameters encompass low-rank LoRA matrices $W_{down}$ and $W_{up}$, prefix-tuning parameters $P_{k}$ and $P_{v}$, adapter parameters, and weights for the gating function. Consequently, UniPELT requires more parameters and inference time than adapter, prefix-tuning, and LoRA, but achieves better performance compared with the performance of the best individual PEFT method.

\subsubsection{Automatic Combination}
Automatic combination explores how to configure PEFT methods like adapters, prefix-tuning, BitFit, and LoRA to different layers of the transformers automatically using various structure search and optimization approaches. However, it typically requires more time and cost due to the need to perform optimization searches in the model or structure. \textbf{AutoPEFT} \cite{zhou2023autopeft} integrates sequential adapter, parallel adapter, and prefix-tuning into the transformer block. The serial adapter receives the hidden state from the FFN output as input, while the parallel adapter takes the hidden state before the FFN layer as its input. In addition, the prefix-tuning module concatenates two prefix vectors, $P_{k}$ and $P_{v}$, with the original key and value vectors, respectively, enabling multi-head attention to adapt to specific target tasks. Motivated by the success of NAS algorithm, AutoPEFT proposes to use the Bayesian optimization approach to automatically search for an appropriate neural architecture network that selectively activates certain layers to incorporate these PEFT modules. Bayesian optimization is not only sample-efficient and zeroth-order but also well-suited for multi-objective setups, enabling cost-efficient optimization and facilitating the trade-off between performance and cost. Moreover, it is more parallelizable during search, which can decrease memory usage. 

\textbf{S$^{3}$Delta-M} (Search for Sparse Structure of Delta Tuning Mix) \cite{hu2022sparse} is a mixture of LoRA, Compacter (low-rank adapter), BitFit, and LNFit\footnote{LNFit is trained only on the variance vectors in the layer normalization module of the PLMs, inspired by \cite{frankle2018the} which trains only on the batch normalization module in convolutional neural networks.}. Different from the simple incorporation of PEFT techniques, S$^{3}$Delta-M is developed by conducting a differentiable delta tuning structure search. It explicitly controls sparsity and searches for an optimal combination of these techniques in a unified search space. In S$^{3}$Delta-M, each PEFT module (LoRA, Compacter, BitFit, and LNFit) is inserted into the corresponding layers of the PLM to ensure the best performance is achieved. The specific combination and placement of these modules are determined through the structure search process, which is guided by explicit sparsity control. \textbf{$\mathcal{S}_{4}$} \cite{chen2023parameterefficient} is a combination of Sequential Adapter, Prefix-tuning, BitFit, and LoRA. Unlike previous methods that utilize the same PEFT module uniformly across all layers of the transformer, $\mathcal{S}_{4}$ is designed by searching for various layer groupings, trainable parameter allocations, tunable groups, and PEFT module assignments. In $\mathcal{S}_{4}$, the layers of the PLMs are divided into four groups, $G_{1}$, $G_{2}$, $G_{3}$, $G_{4}$, in a “spindle” pattern. This means that more layers are allocated to the middle groups ($G_{2}$ and $G_{3}$) while fewer layers are assigned to the top and bottom groups ($G_{1}$ and $G_{4}$). However, all trainable parameters are allocated uniformly, i.e., the number of trainable parameters in each layer remains the same across all groups. Different groups are equipped with different combinations of sequential adapter, prefix-tuning, BitFit, and LoRA. Extensive experimental results demonstrate that better performance is achieved when each group is equipped with the following combinations of PEFT methods. $A$ denotes sequential adapter, $P$ denotes prefix-tuning, $B$ denotes BitFit, and $L$ denotes LoRA.
\begin{gather}
  G_{1} : (A, L); ~~~G_{2} : (A, P);  \notag \\
  G_{3} : (A, P, B); ~~~G_{4} : (P, B, L). \notag 
\end{gather}

\subsection{Unified Fine-tuning}
Unified fine-tuning presents a unified framework for fine-tuning, which streamlines the incorporation of diverse fine-tuning methods into a cohesive architecture, ensuring consistency and efficiency across the adaptation and optimization of models. Unlike hybrid fine-tuning methods, unified fine-tuning methods typically utilize a single PEFT method rather than a combination of various PEFT methods. 

\textbf{AdaMix} \cite{wang-etal-2022-adamix} leverages a mixture of adaptation module approaches to obtain a unified framework for fine-tuning. Motivated by sparsely-activated MoE \cite{shazeer2017outrageously}, AdaMix treats each adaptation module as an individual expert and employs stochastic routing to randomly select a down-projection matrix and an up-projection matrix for weight updates. Such stochastic routing allows the adaption module to learn multiple views for the given task, but it also poses a challenge in deciding which adaption module to use during inference. To this end, Adamix utilizes consistency regularization and adaption module merging (i.e., average weights of all down- and up-projection matrices) to select the trained adaption module and obtain the same computational cost as that of a single module. Notably, adaption modules in Adamix could be adapters like sequential adapter \cite{houlsby2019parameter} or low-rank decomposition matrices like LoRA \cite{hu2022lora}. 

\textbf{SparseAdapter} \cite{he-etal-2022-sparseadapter} utilizes network pruning technique to construct a unified framework in which various PEFT methods, including adapters family and LoRA \cite{houlsby2019parameter,hu2022lora,he2022towards}, can be further pruned to improve parameter efficiency. SparseAdapter sets a target sparsity, denoted as $s$, and assigns a score, denoted as $z$, to all parameters of adapters and LoRA. Parameters with scores below the threshold $z_{s}$ (corresponding to the $s$-th lowest percentile of $z$) are considered redundant and removed. The score $z$ can be computed using pruning methods, such as random pruning, magnitude pruning \cite{frankle2021pruning}, Erdos-Renyi \cite{mocanu2018scalable}, SNIP \cite{lee2018snip}, or GraSP \cite{wang-etal-2018-glue}, based on the adapter weight $W$, with SNIP-based SparseAdapter yielding the best results. Furthermore, SparseAdapter exhibits improved performance compared to full fine-tuning when utilizing the “Large-Sparse” setting, which involves larger bottleneck dimensions and higher sparsity ratios. Notably, the network pruning technique proposed in SparseAdapter is a plug-in method that can be applied to any adapter variants, such as LoRA \cite{hu2022lora}, MAM Adapter \cite{he2022towards}, and AdapterFusion \cite{pfeiffer-etal-2021-adapterfusion}. The optimized parameters in SparseAdapter can be represented as $\hat{W} = W \odot M$, in which $M$ is a binary mask matrix with $M = \mathbb{I}_{\{z \geq z_{s}\}}$ and $z = \text{score}(W)$. 

\textbf{ProPETL} \cite{zeng-etal-2023-one} introduces a single prototype network (e.g., adapter, prefix-tuning, and LoRA) across layers and tasks and constructs different sub-networks for each layer using various binary masks. Inspired by ALBERT \cite{Lan2020ALBERT:}, ProPETL leverages parameter sharing within the prototype network modules in each layer of the transformer, enhancing parameter efficiency and reducing storage requirements. In ProPETL, binary masks $M \in \{0, 1\}^{n}$ are introduced in each layer of the transformer, in which $n$ is the number of parameters in a single PEFT module. Each mask corresponds to a specific sub-network of the shared prototype network. By doing so, though each layer shares the parameters of the same prototype network, each layer has a different sub-network to capture meaningful semantic representations. The final objective of the task adaptation for the PLMs can be expressed as follows:
\begin{align}
\mathop{\text{max}}\limits_{\theta_{pro},m_{1},m_{2},\cdots,m_{L}}\sum_{i=0}^{N}\text{log}P(Y_{i}|X_{i};\theta_{lm},\theta_{sub}), \\ \theta_{sub} = \nonumber [\theta_{pro} \odot m_{1}, \theta_{pro} \odot m_{2}, \cdots, \theta_{pro} \odot m_{L}].
\label{propetl}
\end{align} Here, $\theta_{lm}$ represents the frozen pretrained parameters of the PLMs, $m_{i}$ ($i = 1,2, \cdots, L$) is binary mask matrix, and $\theta_{sub}$ denotes the parameters to be optimized.

\section{Experiments}\label{section4}

%In this section, we select three models from encoder-based, encoder-decoder-based, and decoder-based PLMs and evaluate their fine-tuning performance, parameter efficiency, and memory usage using different PEFT methods.

\subsection{Experimental Settings}
\subsubsection{PLMs and Datasets} 
We use the encoder-only models RoBERTa-base (125M) and RoBERTa-large (355M) \cite{liu2020roberta} to evaluate on the GLUE benchmark \cite{wang-etal-2018-glue}, encoder-decoder models T5-base (220M) and T5-large (770M) \cite{raffel2020exploring} to evaluate on the WMT16 En-Ro dataset\footnote{\url{https://huggingface.co/datasets/wmt16}}, and decoder-only models LLaMA-7B and LLaMA-13B \cite{touvron2023llama} fine-tuned with the Alpaca dataset \cite{taori2023stanford} to evaluate on the MMLU benchmark \cite{hendrycks2021measuring}. All these PLMs with different model types and model scales are based on the encoder, decoder, or encoder-decoder of the \textbf{Transformer} architecture. The datasets we use for experiments cover a wide range of tasks, from NLU to MT and NLG. The GLUE benchmark covers a collection of NLU tasks, including single-sentence classification, and sentence-pair classification tasks. WMT16 En-Ro dataset consists of parallel data pairs, where each pair consists of an English sentence and its corresponding translation into Romanian. Alpaca \cite{taori2023stanford} is an instruction dataset containing 52k samples. MMLU Benchmark \cite{hendrycks2021measuring} encompasses a comprehensive range of 57 disciplines spanning science, humanities, social sciences, and more. The level of difficulty of the benchmark ranges from beginner to advanced levels of expertise, testing both world knowledge and problem-solving abilities.

\subsubsection{PEFT Methods} 
Eleven representative PEFT methods: sequential adapter (Adapter$^{S}$) \cite{houlsby2019parameter}, prompt-tuning \cite{lester-etal-2021-power}, prefix-tuning \cite{li-liang-2021-prefix}, (IA)$^{3}$ \cite{liu2022fewshot}, BitFit \cite{ben-zaken-etal-2022-bitfit}, Child-Tuning \cite{xu-etal-2021-raise}, LoRA \cite{hu2022lora}, AdaLoRA \cite{zhang2023adaptive}, QLoRA \cite{dettmers2023qlora}, MAM adapter \cite{he2022towards}, and ProPELT \cite{zeng-etal-2023-one} are chosen. Since the GLUE benchmark consists of a series of NLU tasks, it serves as the preferred evaluation dataset used by most PLMs to validate the effectiveness of PEFT methods. Ten representative PEFT methods other than QLoRA are selected to fine-tune RoBERTa-base/large. For T5-base/large, we use (IA)$^{3}$ and LoRA for fine-tuning. As for LLaMA-7B/13B, (IA)$^{3}$, LoRA, and QLoRA are used for fine-tuning. 

\subsubsection{Implementation Details} 
Since “prompt-tuning, prefix-tuning, (IA)$^{3}$, LoRA, and AdaLoRA” have been integrated into the PEFT library\footnote{\url{https://huggingface.co/docs/peft/index}}. Therefore, we directly utilize the PEFT library to invoke these PEFT methods for fine-tuning. For BitFit, Child-tuing$_{D}$, MAM adapter, QLoRA, and ProPELT, we experiment using their original code. Significantly, we experiment with sequential adapter using code from the MAM adapter. For RoBERTa-base/large, all PEFT methods are fine-tuned using a batch size of 32 and a sequence length of 128, except for (IA)$^{3}$ which is fine-tuned using batch size 8. We use the batch size 64 for T5-base and 32 for T5-large. For LLaMA-7B/13B, we use batch size 16 for fine-tuning. All experiments are implemented with A800 GPU.

\begin{table*}
\footnotesize
 \centering
 \caption{Fine-tuning RoBERTa-base (RoB$_{B}$) and RoBERTa-large (RoB$_{L}$) models on the GLUE benchmark. Specifically, we report the Matthews correlation for COLA, accuracy/F1 score for MRPC and QQP, Pearson/Spearman correlation for STS-B, averaged matched accuracy for MNLI, and accuracy for other NLU tasks. Higher values indicate better performance across all metrics. We present the number of trainable parameters (\# TPs) of each method, excluding Child-Tuning$_{D}$ due to its randomness during network pruning. We also bold the maximum values and underline the minimum values.}
    \label{glue}
    \renewcommand{\arraystretch}{1.0}
    \begin{tabular}{c|c|c|ccccccccc}
    \hline
    \textbf{Model} & \textbf{PEFT Method} & \textbf{\#TPs} & \textbf{CoLA} & \textbf{SST2} & \textbf{MRPC} & \textbf{STS-B} & \textbf{QQP} & \textbf{MNLI} & \textbf{QNLI} & \textbf{RTE} & \textbf{Avg.} \\
    \hline
    \multirow{11}{*}{RoB$_{B}$} & FT & 124.6M & 59.07 & 92.89 & 88.24/91.58 & 90.87/90.61 & 90.81/87.72 & 86.27 & 91.07 & 72.20 & 84.00/84.00 \\ 
       & Adapter$^{S}$ & 7.41M & 63.32 & 94.31 & \textbf{90.44/93.18} & 91.25/90.94 & 90.81/86.55 &  87.33 & 92.06 & 73.56 & 85.39/85.16 \\ 
        & Prompt-tuning & 0.61M & \underline{49.37} & \underline{92.09} & \underline{70.83/81.72} & \underline{82.44/83.11} & \underline{82.99/78.35} & \underline{80.57} & \underline{80.03} & 58.12 & \underline{74.56/75.42} \\
        & Prefix-tuning & 0.96M & 59.31 & 93.81 & 87.25/91.03 & 88.48/88.32 & 87.75/84.09 & 85.21 & 90.77 & \underline{54.51} & 80.89/80.88\\
       & (IA)$^{3}$ & 0.66M & 59.58 & 93.92 & 87.00/90.52 & 90.30/90.32 & 87.99/84.10 & 83.95 & 90.88 & 71.12 & 83.09/83.05\\
        & BitFit & 0.69M & 61.32 & 94.72 & 89.22/92.41 & 90.34/90.27 & 88.12/84.11 & 84.64 & 91.09 & \textbf{77.98} & 84.68/84.57\\
        & Child-Tuning$_{D}$ & - & 60.33 & 93.58 & 89.22/92.20 & 91.14/90.93 & \textbf{90.98/88.04} & \textbf{87.40} & 92.20 & 77.62 & 85.31/85.29\\
        & LoRA & 0.89M & 62.09 & 94.04 & 87.50/90.68 & 90.66/90.83 & 88.83/85.21 & 86.54 & 92.02 &  72.92 & 84.33/84.29\\
        & AdaLoRA & 1.03M & 59.82 & 93.92 & 87.99/91.33 & 90.83/90.73 & 88.58/84.98 & 86.26 & 91.43 & 70.04 &  83.61/83.56\\
        & MAM Adapter & 46.78M & 61.42 & \textbf{94.87} & 89.31/92.21 & 90.74/90.42 & 88.31/83.20 &  86.63& 90.19  & 72.62 & 84.26/83.95\\
        & ProPELT$_{\text{Adapter}}$ & 1.87M & \textbf{66.33} & 93.85 & 87.25/90.82 & \textbf{91.33/91.04} & 89.22/85.79 & 86.49 & \textbf{92.56} & 75.54 & \textbf{85.32/85.30}\\
        & ProPELT$_{\text{Prefix}}$ & 10.49M & 61.79 & 94.30 & 88.73/91.98 & 90.30/90.19 & 88.54/85.05 & 86.22 & 91.51 & 63.31 & 83.08/83.04\\
        & ProPELT$_{\text{LoRA}}$ & 1.77M & 60.38 & 94.11 & 87.42/90.87 & 90.76/90.55  & 88.90/85.55 & 86.84 & 92.04 & 67.39 & 83.48/83.47\\
    \hline
    \multirow{11}{*}{RoB$_{L}$}  & FT & 355.3M & 65.78 & 95.54 & 89.22/92.28 & 91.74/91.76 & 89.30/86.68 & 89.42 & 93.61 &  81.23 & 86.98/87.04 \\ 
       & Adapter$^{S}$ & 19.77M & 67.03 & 96.37 & 89.94/92.54 & \textbf{92.58/92.42} & \textbf{92.19/88.50} & 91.00 & 94.31 & 85.25 & 88.58/88.43\\ 
        & Prompt-tuning & 1.07M & 61.13 & \underline{94.61} & \underline{73.04/81.29} & \underline{78.51/78.99} & \underline{80.74/75.16} & 68.15 & \underline{89.13} & \underline{60.29} & \underline{75.70/76.09} \\
        & Prefix-tuning & 2.03M & \underline{59.01} & 95.76 & 88.24/91.37 & 90.92/91.07 & 88.88/85.45 & 89.30 & 93.32 & 74.01 & 84.93/84.91\\
       & (IA)$^{3}$ & 1.22M & 61.15 & \underline{94.61} & 86.52/90.33 & 92.22/92.03 & 89.45/86.25 & 88.63 & 94.25 & 81.23 & 86.00/86.06\\
        & BitFit & 1.32M & \textbf{68.01} & 96.10 & \textbf{90.93/93.38} & 91.93/91.77 & 89.48/86.43 & 89.98 & 94.47 & 87.73 & 88.57/88.47\\
        & Child-Tuning$_{D}$ & - &  63.08 &95.07  & 90.69/93.43 & 92.36/92.18 & 91.52/88.75 & \underline{35.45} & 93.15 & 86.25 & 80.95/80.92\\
        & LoRA & 1.84M & 64.47 & \textbf{96.67} & 87.50/91.19 & 91.66/91.44 & 90.15/86.91 & 90.76 & 95.00 & 79.78 & 87.00/87.03\\
        & AdaLoRA & 2.23M & 65.85 & 94.95 & 89.46/92.34 & 92.05/91.80 & 89.60/86.30 & 90.36 & 94.62 & 77.98 & 86.86/86.78\\
        & MAM Adapter & 122.2M &  67.39 & 95.81 & 90.12/92.77 & 92.44/92.18 & 90.87/86.65 & 90.62 & 94.31 & 86.62 & 88.52/88.29\\
        & ProPELT$_{\text{Adapter}}$ & 5.40M & 65.55 & 96.27 & 89.71/92.54 & 91.92/91.67 & 90.67/87.74 & \textbf{91.37} & \textbf{95.20}  & \textbf{88.89} & \textbf{88.70/88.65}\\
        & ProPELT$_{\text{Prefix}}$ & 26.85M & 62.24 & 96.17 & 90.04/92.92  & 90.70/90.49 & 89.30/86.30 & 90.33 & 94.73 & 79.71 & 86.65/86.61\\
        & ProPELT$_{\text{LoRA}}$ & 4.19M & 61.90 & 95.93 & 89.06/92.19 & 91.66/91.38 & 90.93/88.05 & 90.53 & 94.93 & 83.57 & 87.31/87.31\\
    \hline
    \end{tabular}
\end{table*}

% We also report the absolute performance gain over the base LLaMA-7B model and LLaMA-13B model.

\subsection{Fine-tuning Performance and Parameter Efficiency}
\subsubsection{RoBERTa Base/Large on GLUE}
Experimental results of full fine-tuning and 11 representative PEFT methods with RoBERTa-base/large on the GLUE benchmark are presented in Table~\ref{glue}, the following findings are observed:
\begin{itemize}
    \item All PEFT methods reduce the number of trainable parameters, and most PEFT methods achieve performance matching or even better than full fine-tuning on the GLUE benchmark. For RoBERTa-base, the average performance of prompt-tuning, prefix-tuning, IA$^{3}$, AdaLoRA, ProPELT$_{\text{prefix}}$ and ProPELT$_{\text{LoRA}}$ on GLUE all underperforms full finetuning, while that of sequential adapter, BitFit, Child-Tuning$_{D}$, LoRA, MAM adapter, and ProPELT$_{\text{Adapter}}$ outperforms full fine-tuning. For RoBERTa-large, the average performance of prompt-tuning, prefix-tuning, IA3, AdaLoRA, ProPELT$_{\text{prefix}}$ and Child-Tuning$_{D}$ on GLUE underperforms full fine-tuning, while that of sequential adapter, BitFit, LoRA, MAM adapter, ProPELT$_{\text{Adapter}}$ and ProPELT$_{\text{LoRA}}$ outperforms full fine-tuning.
    \item ProPELT$_{\text{adapter}}$, a unified fine-tuning method that employs the AdapterFusion as the backbone, uses about 1.50\% of the trainable parameters to fine-tune RoBERT-base and RoBERTa-large, but achieves optimal average performance on the GLUE benchmark, outperforming RoBERT-base (FT) by about 1.30\% and RoBERT-large (FT) by about 1.65\%.
    \item MAM Adapter, a hybrid fine-tuning method that combines parallel adapters and prefix-tuning, achieves better performance than prefix-tuning, but also consumes a large amount of trainable parameters.
    \item Sequential adapter requires more trainable parameters than prompt-tuning, prefix-tuning, (IA)$^{3}$, BitFit, Child-Tuning$_{D}$, LoRA, and AdaLoRA, but achieves better performance than them on the GLUE benchmark.
    \item Prompt-tuning with the virtual marker length set to 20 achieves the smallest trainable parameter, but also the worst performance, with its average performance on the GLUE benchmark being about 10\% lower than full fine-tuning. 
    \item Child-Tuning$_{D}$ performs well when fine-tuning RoBERT-base on the GLUE benchmark and obtains better performance than full fine-tuning, but performs poorly when fine-tuning RoBERT-large on the MNLI dataset, which we guess it caused by the learning rate.
\end{itemize}
 
%T5 \cite{raffel2020exploring} shows strong performance across a variety of NLP tasks by incorporating task-specific input prefixes. For instance, T5 can be used for translation tasks by adding an appropriate prefix “translate English to Romanian” before the input. 

\subsubsection{T5 Base/Large on WMT16 En-Ro Dataset}
As depicted in Table~\ref{translation}, both (IA)$^{3}$ and LoRA significantly reduce the number of trainable parameters compared to full fine-tuning, while maintaining comparable performance. Specifically, (IA)$^{3}$ employs only 0.03\% of trainable parameters and achieves a BLEU score \cite{papineni2002bleu} 0.16 higher than full fine-tuning for T5-base and 0.01 lower for T5-large. LoRA achieves a BLEU score 0.36 higher than full fine-tuning on T5-base using only 0.39\% of trainable parameters, and 0.01 lower than full fine-tuning on T5-large using only 0.32\% of trainable parameters.

\begin{table}
 \caption{Fine-tuning T5-base and T5-large models on the WMT16 En-Ro dataset and evaluating their performance using BLEU score. The higher BLEU score indicates a better quality of translation output.}
 \centering
    \label{translation}
    \begin{tabular}{c|c|c|c}
    \hline
    \textbf{Model} & \textbf{PEFT Method} & \textbf{\# TPs} & \textbf{BLEU} \\
    \hline
    \multirow{4}{*}{T5-base}  & FT & 222.9M &  27.42 \\ 
       & (IA)$^{3}$ & 0.07M & 27.58 \\
        & LoRA & 0.88M &  \textbf{27.78} \\
    \hline
    \multirow{4}{*}{T5-large}  & FT & 737.7M & \textbf{28.13} \\ 
       & (IA)$^{3}$ & 0.19M &  28.12 \\
        & LoRA & 2.36M & 28.12 \\
    \hline
    \end{tabular}
\end{table}

\subsubsection{LLaMA on MMLU}
\begin{figure*}[t]
\centering
\subfigure[\scriptsize{LLaMA-7B-Alpaca-FT}]{
\includegraphics[width=1.52in]{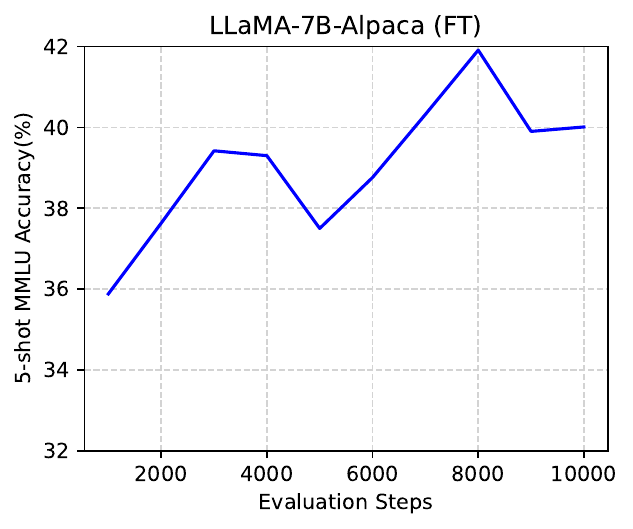}
}
\quad
\subfigure[\scriptsize{LLaMA-7B-Alpaca-(IA)$^{3}$}]{
\includegraphics[width=1.52in]{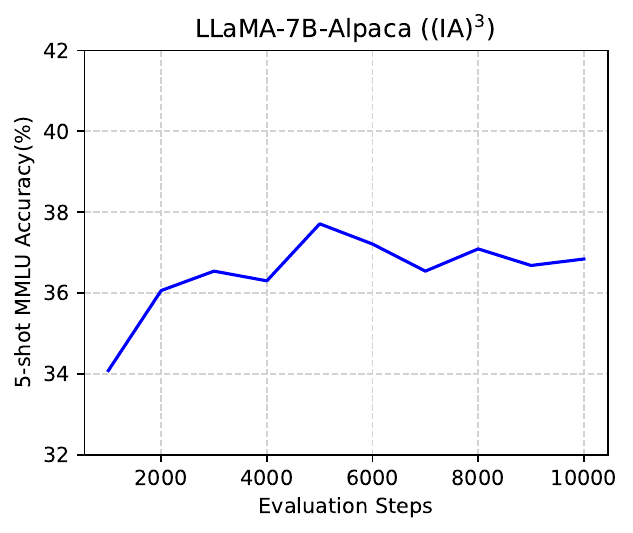}
}
\quad
\subfigure[\scriptsize{LLaMA-7B-Alpaca-LoRA}]{
\includegraphics[width=1.52in]{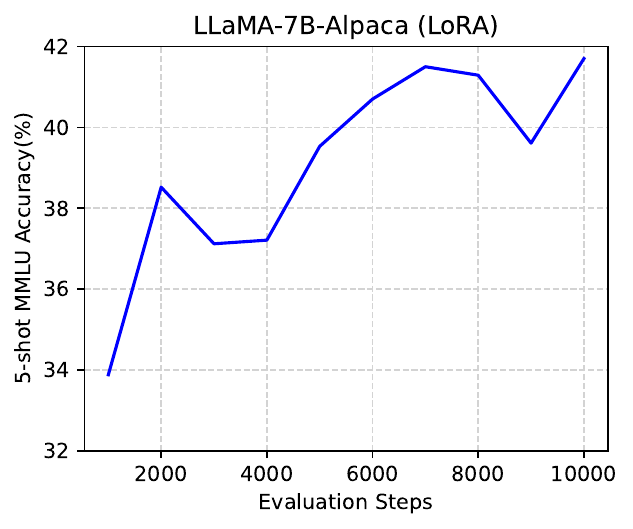}
}
\quad
\centering
\subfigure[\scriptsize{LLaMA-7B-Alpaca-QLoRA}]{
\includegraphics[width=1.52in]{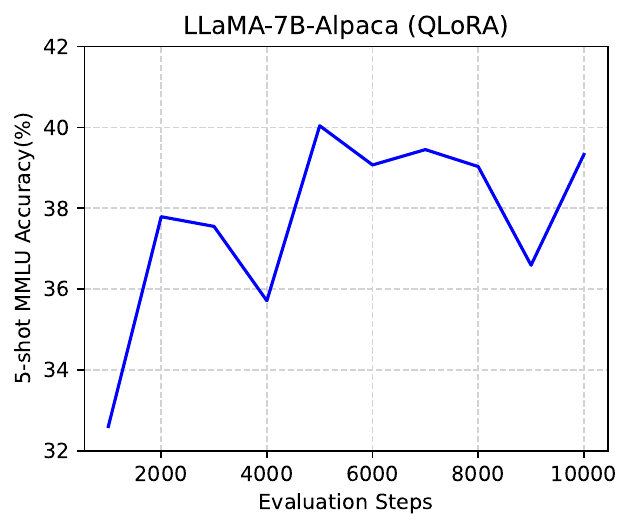}
}
\quad
\subfigure[\scriptsize{LLaMA-13B-Alpaca-FT}]{
\includegraphics[width=1.52in]{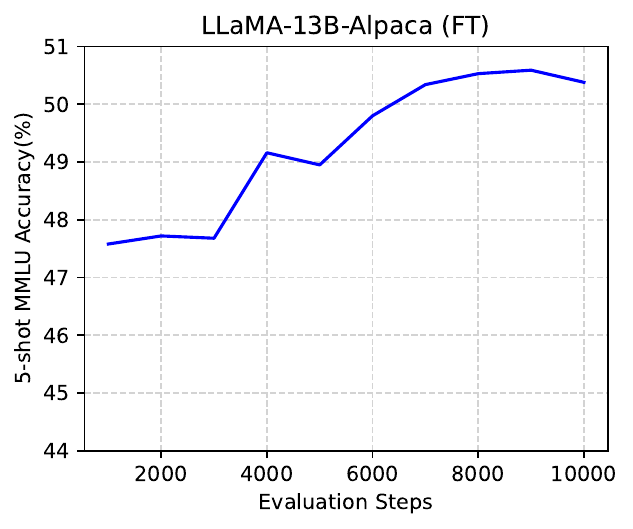}
}
\quad
\subfigure[\scriptsize{LLaMA-13B-Alpaca-(IA)$^{3}$}]{
\includegraphics[width=1.52in]{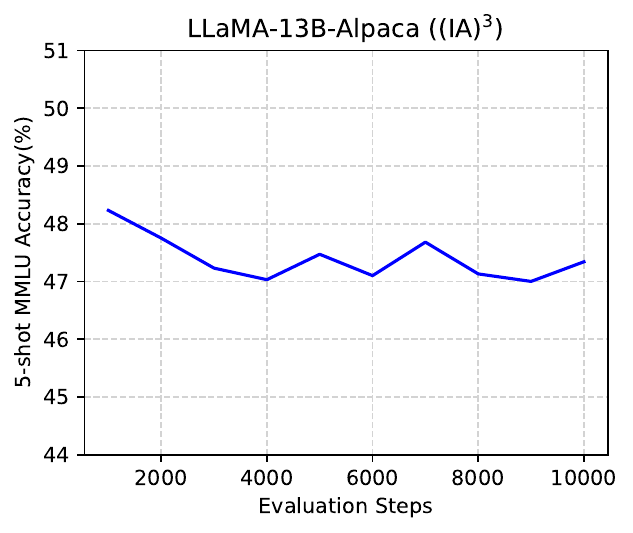}
}
\quad
\subfigure[\scriptsize{LLaMA-13B-Alpaca-LoRA}]{
\includegraphics[width=1.52in]{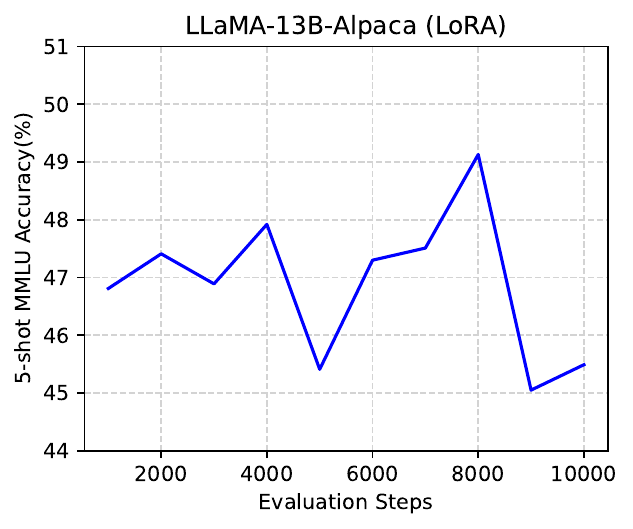}
}
\quad
\centering
\subfigure[\scriptsize{LLaMA-13B-Alpaca-QLoRA}]{
\includegraphics[width=1.52in]{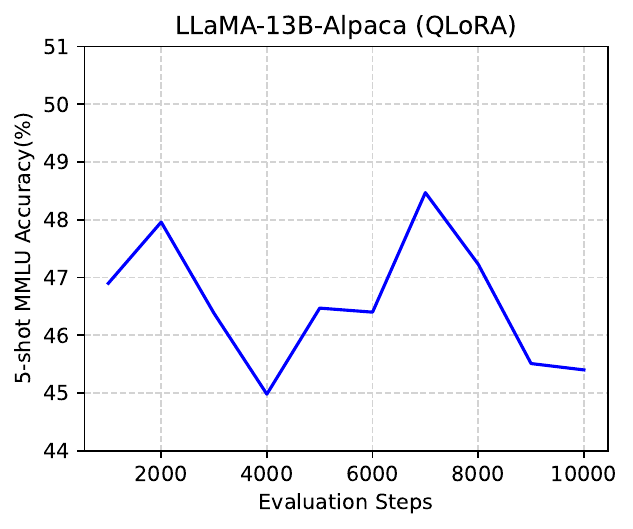}
}
\quad
\caption{The 5-shot accuracy fluctuates on the MMLU dev set with the increase in evaluation steps when fine-tuning LLaMA-7B-Alpaca and LLaMA-7B-Alpaca using the IA$^{3}$, LoRA, and QLoRA methods.}
\label{mmlu_dev}
\end{figure*}

We first tracked the 5-shot MMLU dev accuracy of LLaMA-7B-Alpaca and LLaMA-13B-Alpaca with full fine-tuning and PEFT approaches LoRA, QLoRA, and (IA)$^{3}$, following the work in \cite{dettmers2023qlora}. As depicted in Fig.~\ref{mmlu_dev}, there are significant performance fluctuations in the 5-shot MMLU dev accuracy throughout model training, particularly in LoRA and QLoRA. Moreover, we discovered that full fine-tuning performance of LLaMA-7B-Alpaca on the MMLU benchmark is extremely sensitive to the learning rate, as shown in Table~\ref{difference}. Subsequently, we select the checkpoint with the best performance on the dev set and perform 5-shot accuracy experiments on the test set of the MMLU benchmark. 

\begin{table}[h]
    \caption{Full fine-tuning performance of LLaMA-7B-Alpaca on the test set of MMLU benchmark with different learning rates.}
    \centering
    \renewcommand{\arraystretch}{1.0}
    \begin{tabular}{c|c}
       \hline
        \textbf{Learning rate} & \textbf{5-shot MMLU Accuracy} \\
        \hline
         2e-4 &  25.71   \\
         5e-5 &   26.65  \\
         1e-6 &   \textbf{41.79} \\
         \hline
    \end{tabular}
    \label{difference}
\end{table}

%As shown in Table~\ref{difference}, a slight difference of 4.9e-5 in the learning rate could result in a significant discrepancy of nearly 16\% in the 5-shot MMLU accuracy when full fine-tuning LLaMA-7B with Alpaca dataset.

As illustrated in Table~\ref{mmlu}, full fine-tuning of both LLaMA-7B and LLaMA-13B produces better 5-shot MMLU test accuracy compared to other PEFT methods. (IA)$^{3}$, LoRA, and QLoRA methods all greatly reduce the number of trainable parameters with (IA)$^{3}$ performs best. Although (IA)$^{3}$ only consumes 0.02\% of full fine-tuning parameters, it performs 2-4\% lower than full fine-tuning. LoRA and QLoRA require about 2\% of full fine-tuning parameters, achieving 5-shot MMLU accuracy that is about 2\% lower than full fine-tuning. In particular, QLoRA only uses half the number of trainable parameters of LoRA but achieves comparable performance. This reduction of parameters in QLoRA can be attributed to the incorporation of 4-bit NormalFloat quantization.

\begin{table*}
 \caption{Comparison of the average 5-shot MMLU test accuracy of LLaMA-7B and LLaMA-13B models fine-tuned with Alpaca. The higher the MMLU accuracy, the better. We also report total model parameters (\# APs) and the ratio of trainable parameters.}
 \centering
    \label{mmlu}
    \renewcommand{\arraystretch}{1.0}
    \begin{tabular}{c|c|ccc|c}
    \hline
    \textbf{Model} & \textbf{PEFT Method} & \textbf{\# TPs} & \textbf{\# APs} & \textbf{\% Params} & \textbf{5-shot MMLU Accuracy} \\
    \hline
    \multirow{4}{*}{LLaMA-7B-Alpaca}  & FT & 6738.4M & 6738.4M & 100 & \textbf{41.79} \\ 
         & (IA)$^{3}$ & 1.58M &  6740.0M &  0.02 &  37.88 \\
        & LoRA & 159.9M & 6898.3M & 2.32 &  40.67 \\ 
        & QLoRA & 79.9M & 3660.3M & 2.18 &  39.96\\ 
    \hline
    \multirow{4}{*}{LLaMA-13B-Alpaca}  & FT & 13015.9M & 13015.9M &  100 & \textbf{49.60} \\ 
        & (IA)$^{3}$ & 2.48M &  13018.3M  &  0.02 & 47.42 \\ 
        & LoRA & 250.3M & 13266.2M  & 1.88  & 47.49 \\ 
        & QLoRA & 125.2M &  6922.3M  &  1.81 & 47.29 \\ 
    \hline
    \end{tabular}
\end{table*}

\subsection{Memory Efficiency}
It has been demonstrated that PEFT methods effectively reduce the number of trainable parameters. However, it remains unclear whether they can also reduce GPU memory usage. To assess the impact of PEFT methods on GPU memory, we compare the GPU memory cost of full fine-tuning and PEFT methods across various models and benchmarks. The specific experimental settings can be seen in the section of implementation details. As presented in Table~\ref{memory}, the memory usage of full fine-tuning in RoBERTa, T5, and LLaMA is positively related to total model parameters. RoBERTa, specifically RoBERTa-base, consumes less memory, requiring only 5.38GB. In contrast, LLaMA demands significantly larger memory, notably LLaMA-13B, necessitating approximately 290GB for full fine-tuning.

In RoBERTa-base/large, prompt-tuning, prefix-tuning, (IA)$^{3}$, LoRA and AdaLoRA (implemented using the PEFT library), and BitFit significantly reduce the GPU memory footprint compared to full fine-tuning. Surprisingly, sequential adapter, MAM adapter, Child-Tuning$_{D}$, and ProPELT all use more memory than full fine-tuning. Both sequential adapter and MAM adapter exhibit higher memory consumption, around three times that of full fine-tuning, with the MAM adapter consuming even more memory. For T5-base/large models, (IA)$^{3}$ and LoRA all demonstrate effective memory reduction during fine-tuning, with LoRA outperforming (IA)$^{3}$. Notably, (IA)$^{3}$ consumes less GPU memory than LoRA in RoBERTa-base/large, which is caused by the smaller batch size during (IA)$^{3}$ fine-tuning ((IA)$^{3}$ sets the batch size to 8). 

Likewise, (IA)$^{3}$, LoRA, and QLoRA all significantly reduce the GPU footprint compared to full finetuning in LLaMA-7B/13B. In addition, we discovered that the PEFT method is more effective in reducing memory usage when the number of model parameters is larger. For example, in LLaMA-7B-Alpaca, compared with full fine-tuning, IA$^{3}$, LoRA, and QLoRA reduce memory usage by 24.08\%, 26.30\%, and 66.66\%, respectively; while in LLaMA-13B-Alpaca, compared with full fine-tuning, IA$^{3}$, LoRA, and QLoRA reduce memory usage by 33.55\%, 39.46\% and 76.86\% of memory usage. Notably, QLoRA dramatically reduces GPU memory consumption, with QLoRA fine-tuning the LLaMA-7B requiring only 1/3 of the memory required for full fine-tuning, and fine-tuning the LLaMA-13B requiring less than 1/4 of the memory required for full fine-tuning. This advancement opens up the possibility of fine-tuning LLMs for various downstream tasks in computational resource-constrained scenarios.

\begin{table*}[t]
 \caption{The peak GPU memory usage when fine-tuning RoBERT-base, RoBERTa-large, T5-base, T5-large, LLaMA-7B, and LLaMA-13B model using full fine-tuning and various PEFT methods.}
 \centering
    \label{memory}
    \renewcommand{\arraystretch}{1.0}
    \begin{tabular}{lc | lc }
    \hline
    \textbf{Model \& Method} & \textbf{Memory (GB)} & \textbf{Model \& Method} & \textbf{Memory (GB)}  \\
    \hline
    RoBERTa-base (FT) &   5.38   &   RoBERTa-large (FT) &  11.96  \\
    RoBERTa-base (Adapter$^{S}$) &  15.29  & RoBERTa-large (Adapter$^{S}$) &  37.17 \\
    RoBERTa-base (Prompt-tuning) &  3.84  &   RoBERTa-large (Prompt-tuning) &  7.98  \\
    RoBERTa-base (Prefix-tuning) &  3.56  &   RoBERTa-large (Prefix-tuning) &  7.58  \\
    RoBERTa-base ((IA)$^{3}$) &  2.62  &   RoBERTa-large ((IA)$^{3}$) &  4.83  \\
    RoBERTa-base (BitFit) &  3.27  &   RoBERTa-large (BitFit) &  7.50   \\
    RoBERTa-base (Child-Tuning$_{D}$) & 6.02  &   RoBERTa-large (Child-Tuning$_{D}$) &  13.67  \\
    RoBERTa-base (LoRA) &  3.59  &   RoBERTa-large (LoRA) &  7.50  \\
    RoBERTa-base (AdaLoRA) & 3.57   & RoBERTa-large (AdaLoRA) & 7.43   \\
    RoBERTa-base (MAM Adapter) & \textbf{15.35}  & RoBERTa-large (MAM Adapter) &  \textbf{37.82}  \\
    RoBERTa-base (ProPELT$_{\text{Adapter}}$) &  8.63  &   RoBERTa-large (ProPELT$_{\text{Adapter}}$) &  19.82  \\
    RoBERTa-base (ProPELT$_{\text{Prefix}}$) &  9.47  &   RoBERTa-large (ProPELT$_{\text{Prefix}}$) &  22.85  \\
    RoBERTa-base (ProPELT$_{\text{LoRA}}$) &  8.25  &   RoBERTa-large (ProPELT$_{\text{LoRA}}$) &  19.52  \\
    \hline
    T5-base (FT) &  \textbf{25.17}   &  T5-large (FT)  & \textbf{30.17} \\
    T5-base ((IA)$^{3}$) &  21.36  &   T5-large ((IA)$^{3}$) &  25.71  \\
    T5-base (LoRA) &  19.43  &   T5-large (LoRA) &  23.77  \\
    \hline
    LLaMA-7B-Alpaca (FT)  &  \textbf{169.36}   &  LLaMA-13B-Alpaca (FT)  &  \textbf{287.79}  \\
    LLaMA-7B-Alpaca ((IA)$^{3}$)  &   128.57  &  LLaMA-13B-Alpaca ((IA)$^{3}$)  & 191.24   \\
    LLaMA-7B-Alpaca (LoRA)  &   124.82  &  LLaMA-13B-Alpaca (LoRA)  &  174.24  \\
    LLaMA-7B-Alpaca (QLoRA)  &  56.46   &  LLaMA-13B-Alpaca (QLoRA)  &  66.60   \\
    \hline
    \end{tabular}
\end{table*}

\section{Applications}

%In this section, we present several application scenarios where PEFT methods are used to better illustrate the effectiveness of PEFT, mainly including multi-task learning, cross-lingual transfer, and backdoor attacks and defense.

\subsection{Multi-task Learning}
Multi-task learning is a method that involves training a model on multiple related tasks and exploiting the information shared and transferred between them to improve the performance of each task. PEFT methods such as adapters, prompt-tuning, and LoRA utilize additional modules that can be plugged into PLMs and thus can be used for task-specific fine-tuning to improve generalization of multi-task learning. For instance, studies from \cite{pfeiffer-etal-2021-adapterfusion,karimi-mahabadi-etal-2021-parameter,chronopoulou-etal-2023-adaptersoup,pfeiffer-etal-2020-mad} leverage task-specific adapters to learn information stored in multiple tasks to achieve more robust transfer learning on new tasks. Several works \cite{vu-etal-2022-spot,asai-etal-2022-attempt,wang2023multitask} employ prompt-tuning for multi-task learning. They either utilize pretrained soft prompts from multiple source tasks to initialize the soft prompt of the target task, based on the similarity between the source and target tasks, or employ multi-task data to learn a single shared prompt and transfer it to the target task. Similar to the adapter, a composition of multiple task-specific LoRA modules is also leveraged to transfer knowledge to new tasks \cite{huang2023lorahub, liu2023moelora}. L-LoRA \cite{tang2023parameter} enhances the fusion capabilities of multi-task learning by preventing negative inference between task-specific representations. Additionally, \cite{zhang2023composing} utilizes arithmetic operators, such as the addition and negation operators, to merge parameters of various PEFT methods trained on different tasks for multi-task learning.

\subsection{Cross-Lingual Transfer}
Cross-lingual transfer involves transferring knowledge or models from one language to another. Numerous works have employed PEFT methods, such as adapters, for cross-lingual transfer due to their unique modular design. Bapna and Firat \cite{bapna-firat-2019-simple} utilize sequential adapter \cite{houlsby2019parameter} to fine-tune and restore the performance of a multilingual neural machine translation model on high-resource languages. Artetxe et al. \cite{artetxe-etal-2020-cross} employ sequential adapter \cite{houlsby2019parameter} to transfer a pretrained monolingual model to an unseen language. MAD-X \cite{pfeiffer-etal-2020-mad,pfeiffer-etal-2021-unks} uses language-specific, task-specific, and invertible adapter to learn language-specific and task-specific transformations, as well as address vocabulary mismatches between multilingual and target languages in a modular manner, enabling the adaptation of pretrained multilingual models to target languages. MAD-G \cite{ansell-etal-2021-mad-g} generates language adapters from language representations based on typological features, allowing the sharing of linguistic knowledge across languages for cross-lingual transfer. LT-SFT \cite{ansell-etal-2022-composable} employs sparse fine-tuning to train the model on the source language and learn task-specific sparse difference vectors for cross-lingual transfer. While BAD-X \cite{parovic-etal-2022-bad} trains a bilingual language-pair adapter on both the source and target languages for zero-shot cross-lingual transfer.

\subsection{Backdoor Attacks and Defense}
Backdoor attacks pose a significant security threat, where a small portion of training samples are contaminated with malicious backdoor triggers. When trained on such poisoned datasets, the model behaves normally on benign samples but predicts attacker-selected labels on samples containing the predefined triggers. The susceptibility of PLMs to backdoor attacks poses a substantial risk to real-world applications \cite{tanzer-etal-2022-memorisation}. Building on the vulnerability of pretrained weights, Gu et al. \cite{gu-etal-2023-gradient} employ the PEFT methods to construct backdoor attacks, in which backdoor attacks are directly injected into PEFT modules. However, Zhu et al. \cite{zhu2022moderatefitting} discover that PEFT can serve as a backdoor defense solution by reducing the model capacity via optimizing only a small number of parameters. The findings from \cite{hong2023fewer} also confirm that PEFT can slightly weaken the backdoor attacks and design a novel trojan attack for the PEFT paradigm. 

\section{Further Directions}

%While numerous PEFT methods have emerged, there remain untapped opportunities for further exploration and exploitation in this field. 

\subsection{Lightweight Hybrid PEFT Methods}
There exist many approaches \cite{he2022towards,lawton-etal-2023-neural,karimi2021compacter,mao-etal-2022-unipelt,zhou2023autopeft,hu2022sparse,chen2023parameterefficient} to combine multiple PEFT methods, aiming to leverage the distinctive advantages of each PEFT method and achieve enhanced performance. Nevertheless, the exploration has been limited to PEFT methods such as adapter, LoRA, prefix-tuning, and BitFit, leaving room for further exploitation by incorporating additional combinations of PEFT methods. Moreover, while drawing inspiration from the NAS algorithm, several PEFT methods \cite{zhou2023autopeft,hu2022sparse} have been investigated using diverse optimization techniques to explore optimal neural network architectures for configuring these PEFT methods. There remains potential for continued exploration in utilizing other optimization methods to automatically search for neural network architectures and configure specific combinations of PEFT modules at specific layers. Additionally, utilizing multiple PEFT methods typically results in increased parameter and memory usage, although it enhances performance. Hence, an intriguing research direction involves investigating how to leverage multiple PEFT methods to improve performance while minimizing the number of trainable parameters.

\subsection{LoRA-derived PEFT Methods}
Recently, a multitude of LoRA-based PEFT methods have emerged, as demonstrated in Fig.~\ref{evolution}. These methods further enhance LoRA by incorporating adaptive rank adjustment, unstructured pruning techniques, weight quantization, and multi-task integration. This encourages future research to develop more LoRA-derived PEFT approaches build upon LoRA. Particular emphasis should be placed on pruning technology and weight quantification. The application of pruning techniques can be extended not only to AdaLoRA \cite{zhang2023adaptive} for rank adjustment but also to LoRAPrune \cite{zhang2023pruning} for pruning both pretrained and LoRA weights. Notably, pruning and weight quantization techniques effectively reduce the number of trainable parameters, compress model size, optimize storage and computational requirements of PLMs (especially LLMs), and enhance their utility and scalability across downstream tasks. These techniques can be further explored in conjunction with LoRA to unlock synergistic benefits.

\subsection{Developing PEFT Library}
Numerous PEFT methods have emerged, but employing them is not a straightforward endeavor. To address this challenge, the PEFT library\footnote{\url{https://github.com/huggingface/peft/tree/main}} and AdapterHub\footnote{\url{https://adapterhub.ml/}} have been developed. These libraries integrate commonly used PEFT methods such as prefix-tuning, LoRA, and AdaLoRA. With just a few lines of code, users can directly invoke these PEFT methods, simplifying their usage. Moreover, both the PEFT and AdapterHub libraries offer a range of examples illustrating how to apply these PEFT methods to various PLMs and LLMs for fine-tuning downstream tasks. However, not all PEFT methods are currently integrated into these two libraries. Future efforts can be directed towards expanding the integration of additional methods, further boosting the application development of PEFT methods.

\subsection{Explainability of PEFT Methods}
Though numerous PEFT methods have been proposed, there is a lack of comprehensive studies exploring the reasons behind their ability to achieve comparable performance and reduce trainable parameters. Work from \cite{fu2023effectiveness} unifies PEFT methods under the concept of sparse fine-tuned models and provides a theoretical analysis demonstrating that sparsity can serve as a regularization technique for the original model, effectively controlling the upper bound of stability. While \cite{zeng2023expressive} explores and analyzes the express power of LoRA for fully connected neural networks and transformer networks, showing the conditions under which there exist effective low-rank adapters for a given task. These studies shed light on the working mechanism and effectiveness of certain PEFT methods, but still lack generalization. Future research endeavors could focus on advancing theoretical studies to unravel the underlying working mechanisms of PEFT methods.

\subsection{Exploring PEFT Methods in Computer Vision and Multimodal Learning}
Though PEFT methods have been extensively studied in NLP, their application in computer vision and multimodal learning also shows great potential for further exploration. The sequential adapter in NLP, initially inspired by multi-domain image classification \cite{rebuffi2017learning,rebuffi2018efficient}, has paved the way for rapid advancements in PEFT methods for PLMs. Moreover, researchers have increasingly delved into various PEFT techniques for computer vision \cite{he2023parameter,xu2023bridging}, as well as language-image and image-audio multimodal learning \cite{sung2022vl,pan2022st}, building upon PEFT methods in NLP \cite{houlsby2019parameter,hu2022lora,karimi2021compacter}. However, there is still significant room for further exploration and exploitation in these domains. In particular, PEFT methods hold the potential to facilitate cross-modality transfer in multimodal learning. By fine-tuning pretrained models using PEFT techniques, knowledge acquired from one modality can be effectively transferred to another, resulting in improved performance in multimodal tasks. Consequently, the application of PEFT methods in computer vision and multimodal learning holds tremendous promise as a future research direction.

\section{Conclusions}
This paper presents a comprehensive and structured study of PEFT methods for PLMs. By classifying the PEFT methods in NLP, we identify the main techniques and challenges associated with them. We employ several representative PEFT methods to fine-tune encoder-based RoBERTa, encoder-decoder-based T5, and decoder-based LLaMA on various downstream tasks. Experimental results reveal that most PEFT methods significantly improve parameter efficiency and achieve comparable or even better performance compared to full fine-tuning. Additionally, most PEFT methods lower the memory footprint, with QLoRA drastically reducing the computational memory requirement, and alleviating the memory challenge when fine-tuning LLMs. Furthermore, we introduce common applications of PEFT methods and outline future research directions. As the development of LLMs continues, there is a clear need to develop PEFT methods that can effectively reduce computational resource demands and memory usage during fine-tuning. This survey aims to provide a bird's-eye view of PEFT methods for PLMs and inspiring further research in this area.

\bibliographystyle{IEEEtran} 
\bibliography{New_IEEEtran_how-to}

% Generated by IEEEtran.bst, version: 1.14 (2015/08/26)
\begin{thebibliography}{100}
\providecommand{\url}[1]{#1}
\csname url@samestyle\endcsname
\providecommand{\newblock}{\relax}
\providecommand{\bibinfo}[2]{#2}
\providecommand{\BIBentrySTDinterwordspacing}{\spaceskip=0pt\relax}
\providecommand{\BIBentryALTinterwordstretchfactor}{4}
\providecommand{\BIBentryALTinterwordspacing}{\spaceskip=\fontdimen2\font plus
\BIBentryALTinterwordstretchfactor\fontdimen3\font minus \fontdimen4\font\relax}
\providecommand{\BIBforeignlanguage}[2]{{%
\expandafter\ifx\csname l@#1\endcsname\relax
\typeout{** WARNING: IEEEtran.bst: No hyphenation pattern has been}%
\typeout{** loaded for the language `#1'. Using the pattern for}%
\typeout{** the default language instead.}%
\else
\language=\csname l@#1\endcsname
\fi
#2}}
\providecommand{\BIBdecl}{\relax}
\BIBdecl

\bibitem{devlin-etal-2019-bert}
J.~Devlin, M.-W. Chang, K.~Lee, and K.~Toutanova, ``{BERT}: Pre-training of deep bidirectional transformers for language understanding,'' in \emph{Proc. Conf. North Amer. Chapter Assoc. Comput. Linguistics: Hum. Lang. Technol.}, 2019, pp. 4171--4186.

\bibitem{liu2020roberta}
Y.~Liu, M.~Ott, N.~Goyal, J.~Du, M.~Joshi, D.~Chen, O.~Levy, M.~Lewis, L.~Zettlemoyer, and V.~Stoyanov, ``Roberta: A robustly optimized bert pretraining approach,'' in \emph{Proc. Int. Conf. Learn. Representations}, 2020.

\bibitem{radford2019language}
A.~Radford, J.~Wu, R.~Child, D.~Luan, D.~Amodei, I.~Sutskever \emph{et~al.}, ``Language models are unsupervised multitask learners,'' \emph{OpenAI blog}, vol.~1, no.~8, p.~9, 2019.

\bibitem{raffel2020exploring}
C.~Raffel, N.~Shazeer, A.~Roberts, K.~Lee, S.~Narang, M.~Matena, Y.~Zhou, W.~Li, and P.~J. Liu, ``Exploring the limits of transfer learning with a unified text-to-text transformer,'' \emph{J. Mach. Learn. Res.}, vol.~21, no.~1, pp. 5485--5551, 2020.

\bibitem{zhang2022democratizing}
S.~Zhang, M.~Diab, and L.~Zettlemoyer, ``Democratizing access to large-scale language models with opt-175b,'' \emph{Meta AI}, 2022.

\bibitem{scao2022bloom}
T.~L. Scao, A.~Fan, C.~Akiki, E.~Pavlick, S.~Ili{\'c}, D.~Hesslow, R.~Castagn{\'e}, A.~S. Luccioni, F.~Yvon, M.~Gall{\'e} \emph{et~al.}, ``Bloom: A 176b-parameter open-access multilingual language model,'' \emph{arXiv preprint arXiv:2211.05100}, 2022.

\bibitem{touvron2023llama}
H.~Touvron, T.~Lavril, G.~Izacard, X.~Martinet, M.-A. Lachaux, T.~Lacroix, B.~Rozi{\`e}re, N.~Goyal, E.~Hambro, F.~Azhar \emph{et~al.}, ``Llama: Open and efficient foundation language models,'' \emph{arXiv preprint arXiv:2302.13971}, 2023.

\bibitem{almazrouei2023falcon}
E.~Almazrouei, H.~Alobeidli, A.~Alshamsi, A.~Cappelli, R.~Cojocaru, M.~Alhammadi, M.~Daniele, D.~Heslow, J.~Launay, Q.~Malartic \emph{et~al.}, ``The falcon series of language models: Towards open frontier models,'' \emph{Hugging Face repository}, 2023.

\bibitem{houlsby2019parameter}
N.~Houlsby, A.~Giurgiu, S.~Jastrzebski, B.~Morrone, Q.~De~Laroussilhe, A.~Gesmundo, M.~Attariyan, and S.~Gelly, ``Parameter-efficient transfer learning for nlp,'' in \emph{Proc. Int. Conf. Mach. Learn.}\hskip 1em plus 0.5em minus 0.4em\relax PMLR, 2019, pp. 2790--2799.

\bibitem{li-liang-2021-prefix}
X.~L. Li and P.~Liang, ``Prefix-tuning: Optimizing continuous prompts for generation,'' in \emph{Proc. Annu. Meeting Assoc. Comput. Linguistics, Int. Joint Conf. Natural Lang. Process.}, 2021, pp. 4582--4597.

\bibitem{hu2022lora}
E.~J. Hu, yelong shen, P.~Wallis, Z.~Allen-Zhu, Y.~Li, S.~Wang, L.~Wang, and W.~Chen, ``Lo{RA}: Low-rank adaptation of large language models,'' in \emph{Proc. Int. Conf. Learn. Representations}, 2022.

\bibitem{ding2022delta}
N.~Ding, Y.~Qin, G.~Yang, F.~Wei, Z.~Yang, Y.~Su, S.~Hu, Y.~Chen, C.-M. Chan, W.~Chen \emph{et~al.}, ``Delta tuning: A comprehensive study of parameter efficient methods for pre-trained language models,'' \emph{arXiv preprint arXiv:2203.06904}, 2022.

\bibitem{lialin2023scaling}
V.~Lialin, V.~Deshpande, and A.~Rumshisky, ``Scaling down to scale up: A guide to parameter-efficient fine-tuning,'' \emph{arXiv preprint arXiv:2303.15647}, 2023.

\bibitem{lin-etal-2020-exploring}
Z.~Lin, A.~Madotto, and P.~Fung, ``Exploring versatile generative language model via parameter-efficient transfer learning,'' in \emph{Proc. Findings Conf. Empir. Methods Natural Lang. Process.}, 2020, pp. 441--459.

\bibitem{lei2023conditional}
T.~Lei, J.~Bai, S.~Brahma, J.~Ainslie, K.~Lee, Y.~Zhou, N.~Du, V.~Y. Zhao, Y.~Wu, B.~Li \emph{et~al.}, ``Conditional adapters: Parameter-efficient transfer learning with fast inference,'' \emph{arXiv preprint arXiv:2304.04947}, 2023.

\bibitem{he2022towards}
J.~He, C.~Zhou, X.~Ma, T.~Berg-Kirkpatrick, and G.~Neubig, ``Towards a unified view of parameter-efficient transfer learning,'' in \emph{Proc. Int. Conf. Learn. Representations}, 2022.

\bibitem{ruckle-etal-2021-adapterdrop}
A.~R{\"u}ckl{\'e}, G.~Geigle, M.~Glockner, T.~Beck, J.~Pfeiffer, N.~Reimers, and I.~Gurevych, ``{AdapterDrop}: {O}n the efficiency of adapters in transformers,'' in \emph{Proc. Conf. Empir. Methods Natural Lang. Process.}, 2021, pp. 7930--7946.

\bibitem{zhao-etal-2022-tiny}
H.~Zhao, H.~Tan, and H.~Mei, ``Tiny-attention adapter: Contexts are more important than the number of parameters,'' in \emph{Proc. Conf. Empir. Methods Natural Lang. Process.}, 2022, pp. 6626--6638.

\bibitem{pfeiffer-etal-2021-adapterfusion}
J.~Pfeiffer, A.~Kamath, A.~R{\"u}ckl{\'e}, K.~Cho, and I.~Gurevych, ``{A}dapter{F}usion: Non-destructive task composition for transfer learning,'' in \emph{Proc. Conf. Eur. Chapter Assoc. Comput. Linguistics}, 2021, pp. 487--503.

\bibitem{he2023mera}
S.~He, R.-Z. Fan, L.~Ding, L.~Shen, T.~Zhou, and D.~Tao, ``Mera: Merging pretrained adapters for few-shot learning,'' \emph{arXiv preprint arXiv:2308.15982}, 2023.

\bibitem{karimi-mahabadi-etal-2021-parameter}
R.~Karimi~Mahabadi, S.~Ruder, M.~Dehghani, and J.~Henderson, ``Parameter-efficient multi-task fine-tuning for transformers via shared hypernetworks,'' in \emph{Proc. Annu. Meeting Assoc. Comput. Linguistics, Int. Joint Conf. Natural Lang. Process.}, 2021, pp. 565--576.

\bibitem{chronopoulou-etal-2023-adaptersoup}
A.~Chronopoulou, M.~Peters, A.~Fraser, and J.~Dodge, ``{A}dapter{S}oup: Weight averaging to improve generalization of pretrained language models,'' in \emph{Proc. Findings Assoc. Comput. Linguistics}, 2023, pp. 2054--2063.

\bibitem{hambardzumyan-etal-2021-warp}
K.~Hambardzumyan, H.~Khachatrian, and J.~May, ``{WARP}: {W}ord-level {A}dversarial {R}e{P}rogramming,'' in \emph{Proc. Annu. Meeting Assoc. Comput. Linguistics, Int. Joint Conf. Natural Lang. Process.}, 2021, pp. 4921--4933.

\bibitem{lester-etal-2021-power}
B.~Lester, R.~Al-Rfou, and N.~Constant, ``The power of scale for parameter-efficient prompt tuning,'' in \emph{Proc. Conf. Empir. Methods Natural Lang. Process.}, 2021, pp. 3045--3059.

\bibitem{liu2021gpt}
X.~Liu, Y.~Zheng, Z.~Du, M.~Ding, Y.~Qian, Z.~Yang, and J.~Tang, ``Gpt understands, too,'' \emph{arXiv preprint arXiv:2103.10385}, 2021.

\bibitem{vu-etal-2022-spot}
T.~Vu, B.~Lester, N.~Constant, R.~Al-Rfou{'}, and D.~Cer, ``{SP}o{T}: Better frozen model adaptation through soft prompt transfer,'' in \emph{Proc. Annu. Meeting Assoc. Comput. Linguistics}, 2022, pp. 5039--5059.

\bibitem{asai-etal-2022-attempt}
A.~Asai, M.~Salehi, M.~Peters, and H.~Hajishirzi, ``{ATTEMPT}: Parameter-efficient multi-task tuning via attentional mixtures of soft prompts,'' in \emph{Proc. Conf. Empir. Methods Natural Lang. Process.}, 2022, pp. 6655--6672.

\bibitem{wang2023multitask}
Z.~Wang, R.~Panda, L.~Karlinsky, R.~Feris, H.~Sun, and Y.~Kim, ``Multitask prompt tuning enables parameter-efficient transfer learning,'' in \emph{Proc. Int. Conf. Learn. Representations}, 2023.

\bibitem{sung2022lst}
Y.-L. Sung, J.~Cho, and M.~Bansal, ``{LST}: Ladder side-tuning for parameter and memory efficient transfer learning,'' in \emph{Proc. Adv. Neural Inf. Process. Syst.}, 2022.

\bibitem{liu2022fewshot}
H.~Liu, D.~Tam, M.~Mohammed, J.~Mohta, T.~Huang, M.~Bansal, and C.~Raffel, ``Few-shot parameter-efficient fine-tuning is better and cheaper than in-context learning,'' in \emph{Proc. Adv. Neural Inf. Process. Syst.}, 2022.

\bibitem{yang-etal-2023-parameter}
X.~Yang, J.~Y. Huang, W.~Zhou, and M.~Chen, ``Parameter-efficient tuning with special token adaptation,'' in \emph{Proc. Conf. Eur. Chapter Assoc. Comput. Linguistics}, 2023, pp. 865--872.

\bibitem{cao-etal-2022-attention}
J.~Cao, C.~Satya~Prakash, and W.~Hamza, ``Attention fusion: a light yet efficient late fusion mechanism for task adaptation in {NLU},'' in \emph{Proc. Findings Assoc. Comput. Linguistics}, 2022, pp. 857--866.

\bibitem{chen2023hadamard}
Y.~Chen, Q.~Fu, G.~Fan, L.~Du, J.-G. Lou, S.~Han, D.~Zhang, Z.~Li, and Y.~Xiao, ``Hadamard adapter: An extreme parameter-efficient adapter tuning method for pre-trained language models,'' in \emph{Proc. 32nd ACM Int. Conf. Inf. Knowl. Manage.}, 2023, pp. 276--285.

\bibitem{ben-zaken-etal-2022-bitfit}
E.~Ben~Zaken, Y.~Goldberg, and S.~Ravfogel, ``{B}it{F}it: Simple parameter-efficient fine-tuning for transformer-based masked language-models,'' in \emph{Proc. Annu. Meeting Assoc. Comput. Linguistics}, 2022, pp. 1--9.

\bibitem{lawton-etal-2023-neural}
N.~Lawton, A.~Kumar, G.~Thattai, A.~Galstyan, and G.~Ver~Steeg, ``Neural architecture search for parameter-efficient fine-tuning of large pre-trained language models,'' in \emph{Proc. Findings Assoc. Comput. Linguistics}, 2023, pp. 8506--8515.

\bibitem{zhao-etal-2020-masking}
M.~Zhao, T.~Lin, F.~Mi, M.~Jaggi, and H.~Sch{\"u}tze, ``Masking as an efficient alternative to finetuning for pretrained language models,'' in \emph{Proc. Conf. Empir. Methods Natural Lang. Process.}, 2020, pp. 2226--2241.

\bibitem{sung2021training}
Y.-L. Sung, V.~Nair, and C.~Raffel, ``Training neural networks with fixed sparse masks,'' in \emph{Proc. Adv. Neural Inf. Process. Syst.}, 2021.

\bibitem{ansell-etal-2022-composable}
A.~Ansell, E.~Ponti, A.~Korhonen, and I.~Vuli{\'c}, ``Composable sparse fine-tuning for cross-lingual transfer,'' in \emph{Proc. Annu. Meeting Assoc. Comput. Linguistics}, 2022, pp. 1778--1796.

\bibitem{xu-etal-2021-raise}
R.~Xu, F.~Luo, Z.~Zhang, C.~Tan, B.~Chang, S.~Huang, and F.~Huang, ``Raise a child in large language model: Towards effective and generalizable fine-tuning,'' in \emph{Proc. Conf. Empir. Methods Natural Lang. Process.}, 2021, pp. 9514--9528.

\bibitem{guo-etal-2021-parameter}
D.~Guo, A.~Rush, and Y.~Kim, ``Parameter-efficient transfer learning with diff pruning,'' in \emph{Proc. Annu. Meeting Assoc. Comput. Linguistics, Int. Joint Conf. Natural Lang. Process.}, 2021, pp. 4884--4896.

\bibitem{fu2023effectiveness}
Z.~Fu, H.~Yang, A.~M.-C. So, W.~Lam, L.~Bing, and N.~Collier, ``On the effectiveness of parameter-efficient fine-tuning,'' in \emph{Proc. AAAI Conf. Artif. Intell.}, vol.~37, no.~11, 2023, pp. 12\,799--12\,807.

\bibitem{aghajanyan-etal-2021-intrinsic}
A.~Aghajanyan, S.~Gupta, and L.~Zettlemoyer, ``Intrinsic dimensionality explains the effectiveness of language model fine-tuning,'' in \emph{Proc. Annu. Meeting Assoc. Comput. Linguistics, Int. Joint Conf. Natural Lang. Process.}, 2021, pp. 7319--7328.

\bibitem{edalati2022krona}
A.~Edalati, M.~Tahaei, I.~Kobyzev, V.~P. Nia, J.~J. Clark, and M.~Rezagholizadeh, ``Krona: Parameter efficient tuning with kronecker adapter,'' \emph{arXiv preprint arXiv:2212.10650}, 2022.

\bibitem{valipour-etal-2023-dylora}
M.~Valipour, M.~Rezagholizadeh, I.~Kobyzev, and A.~Ghodsi, ``{D}y{L}o{RA}: Parameter-efficient tuning of pre-trained models using dynamic search-free low-rank adaptation,'' in \emph{Proc. Conf. Eur. Chapter Assoc. Comput. Linguistics}, 2023, pp. 3274--3287.

\bibitem{zhang2023adaptive}
Q.~Zhang, M.~Chen, A.~Bukharin, P.~He, Y.~Cheng, W.~Chen, and T.~Zhao, ``Adaptive budget allocation for parameter-efficient fine-tuning,'' in \emph{Proc. Int. Conf. Learn. Representations}, 2023.

\bibitem{zhang2023increlora}
F.~Zhang, L.~Li, J.~Chen, Z.~Jiang, B.~Wang, and Y.~Qian, ``Increlora: Incremental parameter allocation method for parameter-efficient fine-tuning,'' \emph{arXiv preprint arXiv:2308.12043}, 2023.

\bibitem{zi2023delta}
B.~Zi, X.~Qi, L.~Wang, J.~Wang, K.-F. Wong, and L.~Zhang, ``Delta-lora: Fine-tuning high-rank parameters with the delta of low-rank matrices,'' \emph{arXiv preprint arXiv:2309.02411}, 2023.

\bibitem{zhang2023pruning}
M.~Zhang, C.~Shen, Z.~Yang, L.~Ou, X.~Yu, B.~Zhuang \emph{et~al.}, ``Pruning meets low-rank parameter-efficient fine-tuning,'' \emph{arXiv preprint arXiv:2305.18403}, 2023.

\bibitem{dettmers2023qlora}
T.~Dettmers, A.~Pagnoni, A.~Holtzman, and L.~Zettlemoyer, ``Qlora: Efficient finetuning of quantized llms,'' \emph{arXiv preprint arXiv:2305.14314}, 2023.

\bibitem{xu2023qa}
Y.~Xu, L.~Xie, X.~Gu, X.~Chen, H.~Chang, H.~Zhang, Z.~Chen, X.~Zhang, and Q.~Tian, ``Qa-lora: Quantization-aware low-rank adaptation of large language models,'' \emph{arXiv preprint arXiv:2309.14717}, 2023.

\bibitem{li2023loftq}
Y.~Li, Y.~Yu, C.~Liang, P.~He, N.~Karampatziakis, W.~Chen, and T.~Zhao, ``Loftq: Lora-fine-tuning-aware quantization for large language models,'' \emph{arXiv preprint arXiv:2310.08659}, 2023.

\bibitem{chen-etal-2022-empowering}
Y.~Chen, D.~Hazarika, M.~Namazifar, Y.~Liu, D.~Jin, and D.~Hakkani-Tur, ``Empowering parameter-efficient transfer learning by recognizing the kernel structure in self-attention,'' in \emph{Proc. Findings Assoc. Comput. Linguistics}, 2022, pp. 1375--1388.

\bibitem{yang2023bayesian}
A.~X. Yang, M.~Robeyns, X.~Wang, and L.~Aitchison, ``Bayesian low-rank adaptation for large language models,'' \emph{arXiv preprint arXiv:2308.13111}, 2023.

\bibitem{zhang2023lora}
L.~Zhang, L.~Zhang, S.~Shi, X.~Chu, and B.~Li, ``Lora-fa: Memory-efficient low-rank adaptation for large language models fine-tuning,'' \emph{arXiv preprint arXiv:2308.03303}, 2023.

\bibitem{huang2023lorahub}
C.~Huang, Q.~Liu, B.~Y. Lin, T.~Pang, C.~Du, and M.~Lin, ``Lorahub: Efficient cross-task generalization via dynamic lora composition,'' \emph{arXiv preprint arXiv:2307.13269}, 2023.

\bibitem{liu2023moelora}
Q.~Liu, X.~Wu, X.~Zhao, Y.~Zhu, D.~Xu, F.~Tian, and Y.~Zheng, ``Moelora: An moe-based parameter efficient fine-tuning method for multi-task medical applications,'' \emph{arXiv preprint arXiv:2310.18339}, 2023.

\bibitem{tang2023parameter}
A.~Tang, L.~Shen, Y.~Luo, Y.~Zhan, H.~Hu, B.~Du, Y.~Chen, and D.~Tao, ``Parameter efficient multi-task model fusion with partial linearization,'' \emph{arXiv preprint arXiv:2310.04742}, 2023.

\bibitem{karimi2021compacter}
R.~Karimi~Mahabadi, J.~Henderson, and S.~Ruder, ``Compacter: Efficient low-rank hypercomplex adapter layers,'' \emph{Proc. Adv. Neural Inf. Process. Syst.}, vol.~34, pp. 1022--1035, 2021.

\bibitem{mao-etal-2022-unipelt}
Y.~Mao, L.~Mathias, R.~Hou, A.~Almahairi, H.~Ma, J.~Han, S.~Yih, and M.~Khabsa, ``{U}ni{PELT}: A unified framework for parameter-efficient language model tuning,'' in \emph{Proc. Annu. Meeting Assoc. Comput. Linguistics}, 2022, pp. 6253--6264.

\bibitem{zhou2023autopeft}
H.~Zhou, X.~Wan, I.~Vuli{\'c}, and A.~Korhonen, ``Autopeft: Automatic configuration search for parameter-efficient fine-tuning,'' \emph{arXiv preprint arXiv:2301.12132}, 2023.

\bibitem{hu2022sparse}
S.~Hu, Z.~Zhang, N.~Ding, Y.~Wang, Y.~Wang, Z.~Liu, and M.~Sun, ``Sparse structure search for delta tuning,'' \emph{Proc. Adv. Neural Inf. Process. Syst.}, vol.~35, pp. 9853--9865, 2022.

\bibitem{chen2023parameterefficient}
J.~Chen, A.~Zhang, X.~Shi, M.~Li, A.~Smola, and D.~Yang, ``Parameter-efficient fine-tuning design spaces,'' in \emph{Proc. Int. Conf. Learn. Representations}, 2023.

\bibitem{wang-etal-2022-adamix}
Y.~Wang, S.~Agarwal, S.~Mukherjee, X.~Liu, J.~Gao, A.~H. Awadallah, and J.~Gao, ``{A}da{M}ix: Mixture-of-adaptations for parameter-efficient model tuning,'' in \emph{Proc. Conf. Empir. Methods Natural Lang. Process.}, 2022, pp. 5744--5760.

\bibitem{he-etal-2022-sparseadapter}
S.~He, L.~Ding, D.~Dong, J.~Zhang, and D.~Tao, ``{S}parse{A}dapter: An easy approach for improving the parameter-efficiency of adapters,'' in \emph{Proc. Findings Conf. Empir. Methods Natural Lang. Process.}, 2022, pp. 2184--2190.

\bibitem{zeng-etal-2023-one}
G.~Zeng, P.~Zhang, and W.~Lu, ``One network, many masks: Towards more parameter-efficient transfer learning,'' in \emph{Proc. Annu. Meeting Assoc. Comput. Linguistics}, 2023, pp. 7564--7580.

\bibitem{vaswani2017attention}
A.~Vaswani, N.~Shazeer, N.~Parmar, J.~Uszkoreit, L.~Jones, A.~N. Gomez, {\L}.~Kaiser, and I.~Polosukhin, ``Attention is all you need,'' \emph{Proc. Adv. Neural Inf. Process. Syst.}, vol.~30, 2017.

\bibitem{he2016deep}
K.~He, X.~Zhang, S.~Ren, and J.~Sun, ``Deep residual learning for image recognition,'' in \emph{Proc. IEEE Conf. Comput. Vis. Pattern Recognit.}, 2016, pp. 770--778.

\bibitem{ba2016layer}
J.~L. Ba, J.~R. Kiros, and G.~E. Hinton, ``Layer normalization,'' \emph{arXiv preprint arXiv:1607.06450}, 2016.

\bibitem{xu2023improving}
L.~Xu and W.~Wang, ``Improving aspect-based sentiment analysis with contrastive learning,'' \emph{Natural Language Processing Journal}, vol.~3, p. 100009, 2023.

\bibitem{xie2020distant}
Y.~Xie, W.~Yang, L.~Tan, K.~Xiong, N.~J. Yuan, B.~Huai, M.~Li, and J.~Lin, ``Distant supervision for multi-stage fine-tuning in retrieval-based question answering,'' in \emph{Proceedings of The Web Conference}, 2020, pp. 2934--2940.

\bibitem{dabre2019exploiting}
R.~Dabre, A.~Fujita, and C.~Chu, ``Exploiting multilingualism through multistage fine-tuning for low-resource neural machine translation,'' in \emph{Proc. Conf. Empir. Methods Natural Lang. Process., Int. Joint Conf. Natural Lang. Process.}, 2019, pp. 1410--1416.

\bibitem{hosseini2023towards}
M.~T. Hosseini, A.~Ghaffari, M.~S. Tahaei, M.~Rezagholizadeh, M.~Asgharian, and V.~P. Nia, ``Towards fine-tuning pre-trained language models with integer forward and backward propagation,'' in \emph{Proc. Findings Assoc. Comput. Linguistics}, 2023, pp. 1867--1876.

\bibitem{amari1993backpropagation}
S.-i. Amari, ``Backpropagation and stochastic gradient descent method,'' \emph{Neurocomputing}, vol.~5, no. 4-5, pp. 185--196, 1993.

\bibitem{xu2023contrastive}
L.~Xu, H.~Xie, Z.~Li, F.~L. Wang, W.~Wang, and Q.~Li, ``Contrastive learning models for sentence representations,'' \emph{ACM Trans. Intel. Syst. Tec.}, vol.~14, no.~4, pp. 1--34, 2023.

\bibitem{pfeiffer-etal-2020-mad}
J.~Pfeiffer, I.~Vuli{\'c}, I.~Gurevych, and S.~Ruder, ``{MAD-X}: {A}n {A}dapter-{B}ased {F}ramework for {M}ulti-{T}ask {C}ross-{L}ingual {T}ransfer,'' in \emph{Proc. Conf. Empir. Methods Natural Lang. Process.}, 2020, pp. 7654--7673.

\bibitem{zhu2021counter}
Y.~Zhu, J.~Feng, C.~Zhao, M.~Wang, and L.~Li, ``Counter-interference adapter for multilingual machine translation,'' \emph{arXiv preprint arXiv:2104.08154}, 2021.

\bibitem{rebuffi2017learning}
S.-A. Rebuffi, H.~Bilen, and A.~Vedaldi, ``Learning multiple visual domains with residual adapters,'' \emph{Proc. Adv. Neural Inf. Process. Syst.}, vol.~30, 2017.

\bibitem{solomon2015convolutional}
J.~Solomon, F.~De~Goes, G.~Peyr{\'e}, M.~Cuturi, A.~Butscher, A.~Nguyen, T.~Du, and L.~Guibas, ``Convolutional wasserstein distances: Efficient optimal transportation on geometric domains,'' \emph{ACM Trans. Graph.}, vol.~34, no.~4, pp. 1--11, 2015.

\bibitem{singh2020model}
S.~P. Singh and M.~Jaggi, ``Model fusion via optimal transport,'' \emph{Proc. Adv. Neural Inf. Process. Syst.}, vol.~33, pp. 22\,045--22\,055, 2020.

\bibitem{ha2017hypernetworks}
D.~Ha, A.~M. Dai, and Q.~V. Le, ``Hypernetworks,'' in \emph{Proc. Int. Conf. Learn. Representations}, 2017.

\bibitem{aharoni-goldberg-2020-unsupervised}
R.~Aharoni and Y.~Goldberg, ``Unsupervised domain clusters in pretrained language models,'' in \emph{Proc. Annu. Meeting Assoc. Comput. Linguistics}, 2020, pp. 7747--7763.

\bibitem{li2017pruning}
H.~Li, A.~Kadav, I.~Durdanovic, H.~Samet, and H.~P. Graf, ``Pruning filters for efficient convnets,'' in \emph{Proc. Int. Conf. Learn. Representations}, 2017.

\bibitem{clark-etal-2019-bert}
K.~Clark, U.~Khandelwal, O.~Levy, and C.~D. Manning, ``What does {BERT} look at? an analysis of {BERT}{'}s attention,'' in \emph{Proc. of 2019 ACL Workshop BlackboxNLP}, 2019, pp. 276--286.

\bibitem{kovaleva-etal-2019-revealing}
O.~Kovaleva, A.~Romanov, A.~Rogers, and A.~Rumshisky, ``Revealing the dark secrets of {BERT},'' in \emph{Proc. Conf. Empir. Methods Natural Lang. Process., Int. Joint Conf. Natural Lang. Process.}, 2019, pp. 4365--4374.

\bibitem{elsken2019neural}
T.~Elsken, J.~H. Metzen, and F.~Hutter, ``Neural architecture search: A survey,'' \emph{J. Mach. Learn. Res.}, vol.~20, no.~1, pp. 1997--2017, 2019.

\bibitem{frankle2018the}
J.~Frankle and M.~Carbin, ``The lottery ticket hypothesis: Finding sparse, trainable neural networks,'' in \emph{Proc. Int. Conf. Learn. Representations}, 2019.

\bibitem{le2013fastfood}
Q.~Le, T.~Sarl{\'o}s, and A.~Smola, ``Fastfood-computing hilbert space expansions in loglinear time,'' in \emph{Proc. Int. Conf. Mach. Learn.}\hskip 1em plus 0.5em minus 0.4em\relax PMLR, 2013, pp. 244--252.

\bibitem{molchanov2019importance}
P.~Molchanov, A.~Mallya, S.~Tyree, I.~Frosio, and J.~Kautz, ``Importance estimation for neural network pruning,'' in \emph{Proc. IEEE Conf. Comput. Vis. Pattern Recognit.}, 2019, pp. 11\,264--11\,272.

\bibitem{sanh2020movement}
V.~Sanh, T.~Wolf, and A.~Rush, ``Movement pruning: Adaptive sparsity by fine-tuning,'' \emph{Proc. Adv. Neural Inf. Process. Syst.}, vol.~33, pp. 20\,378--20\,389, 2020.

\bibitem{mackay1992practical}
D.~J. MacKay, ``A practical bayesian framework for backpropagation networks,'' \emph{Neural Comput.}, vol.~4, no.~3, pp. 448--472, 1992.

\bibitem{liu2020versatile}
J.~Liu, A.~Moreau, M.~Preuss, J.~Rapin, B.~Roziere, F.~Teytaud, and O.~Teytaud, ``Versatile black-box optimization,'' in \emph{Proc. of the 2020 Genet. and Evolut. Comput. Conf.}, 2020, pp. 620--628.

\bibitem{ilharco2023editing}
G.~Ilharco, M.~T. Ribeiro, M.~Wortsman, L.~Schmidt, H.~Hajishirzi, and A.~Farhadi, ``Editing models with task arithmetic,'' in \emph{Proc. Int. Conf. Learn. Representations}, 2023.

\bibitem{zhang2023composing}
J.~Zhang, S.~Chen, J.~Liu, and J.~He, ``Composing parameter-efficient modules with arithmetic operations,'' \emph{arXiv preprint arXiv:2306.14870}, 2023.

\bibitem{yadav2023resolving}
P.~Yadav, D.~Tam, L.~Choshen, C.~Raffel, and M.~Bansal, ``Resolving interference when merging models,'' \emph{arXiv preprint arXiv:2306.01708}, 2023.

\bibitem{zhang2021beyond}
A.~Zhang, Y.~Tay, S.~Zhang, A.~Chan, A.~T. Luu, S.~Hui, and J.~Fu, ``Beyond fully-connected layers with quaternions: Parameterization of hypercomplex multiplications with \$1/n\$ parameters,'' in \emph{Proc. Int. Conf. Learn. Representations}, 2021.

\bibitem{shazeer2017outrageously}
N.~Shazeer, A.~Mirhoseini, K.~Maziarz, A.~Davis, Q.~Le, G.~Hinton, and J.~Dean, ``Outrageously large neural networks: The sparsely-gated mixture-of-experts layer,'' \emph{arXiv preprint arXiv:1701.06538}, 2017.

\bibitem{frankle2021pruning}
J.~Frankle, G.~K. Dziugaite, D.~Roy, and M.~Carbin, ``Pruning neural networks at initialization: Why are we missing the mark?'' in \emph{Proc. Int. Conf. Learn. Representations}, 2021.

\bibitem{mocanu2018scalable}
D.~C. Mocanu, E.~Mocanu, P.~Stone, P.~H. Nguyen, M.~Gibescu, and A.~Liotta, ``Scalable training of artificial neural networks with adaptive sparse connectivity inspired by network science,'' \emph{Nature communications}, vol.~9, no.~1, p. 2383, 2018.

\bibitem{lee2018snip}
N.~Lee, T.~Ajanthan, and P.~Torr, ``{SNIP}: Single-shot network pruning based on connection sensitivity,'' in \emph{Proc. Int. Conf. Learn. Representations}, 2019.

\bibitem{wang-etal-2018-glue}
A.~Wang, A.~Singh, J.~Michael, F.~Hill, O.~Levy, and S.~Bowman, ``{GLUE}: A multi-task benchmark and analysis platform for natural language understanding,'' in \emph{Proc. of 2018 EMNLP Workshop BlackboxNLP}, 2018, pp. 353--355.

\bibitem{Lan2020ALBERT:}
Z.~Lan, M.~Chen, S.~Goodman, K.~Gimpel, P.~Sharma, and R.~Soricut, ``Albert: A lite bert for self-supervised learning of language representations,'' in \emph{Proc. Int. Conf. Learn. Representations}, 2020.

\bibitem{taori2023stanford}
R.~Taori, I.~Gulrajani, T.~Zhang, Y.~Dubois, X.~Li, C.~Guestrin, P.~Liang, and T.~B. Hashimoto, ``Stanford alpaca: An instruction-following llama model,'' 2023.

\bibitem{hendrycks2021measuring}
D.~Hendrycks, C.~Burns, S.~Basart, A.~Zou, M.~Mazeika, D.~Song, and J.~Steinhardt, ``Measuring massive multitask language understanding,'' in \emph{Proc. Int. Conf. Learn. Representations}, 2021.

\bibitem{papineni2002bleu}
K.~Papineni, S.~Roukos, T.~Ward, and W.-J. Zhu, ``Bleu: a method for automatic evaluation of machine translation,'' in \emph{Proc. Annu. Meeting Assoc. Comput. Linguistics}, 2002, pp. 311--318.

\bibitem{bapna-firat-2019-simple}
A.~Bapna and O.~Firat, ``Simple, scalable adaptation for neural machine translation,'' in \emph{Proc. Conf. Empir. Methods Natural Lang. Process., Int. Joint Conf. Natural Lang. Process.}, 2019, pp. 1538--1548.

\bibitem{artetxe-etal-2020-cross}
M.~Artetxe, S.~Ruder, and D.~Yogatama, ``On the cross-lingual transferability of monolingual representations,'' in \emph{Proc. Annu. Meeting Assoc. Comput. Linguistics}, 2020, pp. 4623--4637.

\bibitem{pfeiffer-etal-2021-unks}
J.~Pfeiffer, I.~Vuli{\'c}, I.~Gurevych, and S.~Ruder, ``{UNK}s everywhere: {A}dapting multilingual language models to new scripts,'' in \emph{Proc. Conf. Empir. Methods Natural Lang. Process.}, 2021, pp. 10\,186--10\,203.

\bibitem{ansell-etal-2021-mad-g}
A.~Ansell, E.~M. Ponti, J.~Pfeiffer, S.~Ruder, G.~Glava{\v{s}}, I.~Vuli{\'c}, and A.~Korhonen, ``{MAD}-{G}: {M}ultilingual adapter generation for efficient cross-lingual transfer,'' in \emph{Proc. Findings Conf. Empir. Methods Natural Lang. Process.}, 2021, pp. 4762--4781.

\bibitem{parovic-etal-2022-bad}
M.~Parovi{\'c}, G.~Glava{\v{s}}, I.~Vuli{\'c}, and A.~Korhonen, ``{BAD}-{X}: Bilingual adapters improve zero-shot cross-lingual transfer,'' in \emph{Proc. Conf. North Amer. Chapter Assoc. Comput. Linguistics: Hum. Lang. Technol.}, 2022, pp. 1791--1799.

\bibitem{tanzer-etal-2022-memorisation}
M.~T{\"a}nzer, S.~Ruder, and M.~Rei, ``Memorisation versus generalisation in pre-trained language models,'' in \emph{Proc. Annu. Meeting Assoc. Comput. Linguistics}, 2022, pp. 7564--7578.

\bibitem{gu-etal-2023-gradient}
N.~Gu, P.~Fu, X.~Liu, Z.~Liu, Z.~Lin, and W.~Wang, ``A gradient control method for backdoor attacks on parameter-efficient tuning,'' in \emph{Proc. Annu. Meeting Assoc. Comput. Linguistics}, 2023, pp. 3508--3520.

\bibitem{zhu2022moderatefitting}
B.~Zhu, Y.~Qin, G.~Cui, Y.~Chen, W.~Zhao, C.~Fu, Y.~Deng, Z.~Liu, J.~Wang, W.~Wu, M.~Sun, and M.~Gu, ``Moderate-fitting as a natural backdoor defender for pre-trained language models,'' in \emph{Proc. Adv. Neural Inf. Process. Syst.}, 2022.

\bibitem{hong2023fewer}
L.~Hong and T.~Wang, ``Fewer is more: Trojan attacks on parameter-efficient fine-tuning,'' \emph{arXiv preprint arXiv:2310.00648}, 2023.

\bibitem{zeng2023expressive}
Y.~Zeng and K.~Lee, ``The expressive power of low-rank adaptation,'' \emph{arXiv preprint arXiv:2310.17513}, 2023.

\bibitem{rebuffi2018efficient}
S.-A. Rebuffi, H.~Bilen, and A.~Vedaldi, ``Efficient parametrization of multi-domain deep neural networks,'' in \emph{Proc. IEEE Conf. Comput. Vis. Pattern Recognit.}, 2018, pp. 8119--8127.

\bibitem{he2023parameter}
X.~He, C.~Li, P.~Zhang, J.~Yang, and X.~E. Wang, ``Parameter-efficient model adaptation for vision transformers,'' in \emph{Proc. AAAI Conf. Artif. Intell.}, vol.~37, no.~1, 2023, pp. 817--825.

\bibitem{xu2023bridging}
Z.~Xu, Z.~Chen, Y.~Zhang, Y.~Song, X.~Wan, and G.~Li, ``Bridging vision and language encoders: Parameter-efficient tuning for referring image segmentation,'' in \emph{IEEE Int. Conf. Comput. Vis.}, 2023, pp. 17\,503--17\,512.

\bibitem{sung2022vl}
Y.-L. Sung, J.~Cho, and M.~Bansal, ``Vl-adapter: Parameter-efficient transfer learning for vision-and-language tasks,'' in \emph{Proc. IEEE Conf. Comput. Vis. Pattern Recognit.}, 2022, pp. 5227--5237.

\bibitem{pan2022st}
J.~Pan, Z.~Lin, X.~Zhu, J.~Shao, and H.~Li, ``St-adapter: Parameter-efficient image-to-video transfer learning,'' \emph{Proc. Adv. Neural Inf. Process. Syst.}, vol.~35, pp. 26\,462--26\,477, 2022.

\end{thebibliography}

\begin{IEEEbiography}[{\includegraphics[width=1in,height=1.25in,clip,keepaspectratio]{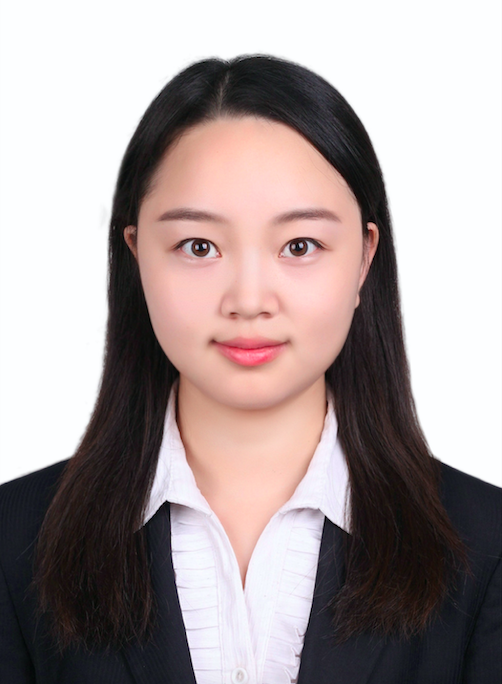}}]{Lingling Xu} (Student Member, IEEE) is currently pursuing her Ph.D. degree at Hong Kong Metropolitan University. She received a Master degree in Mathematics from Shandong University. Her research interests include parameter-efficient fine-tuning, contrastive learning, representation learning, and aspect-based sentiment analysis.\end{IEEEbiography}

\begin{IEEEbiography}[{\includegraphics[width=1in,height=1.25in,clip,keepaspectratio]{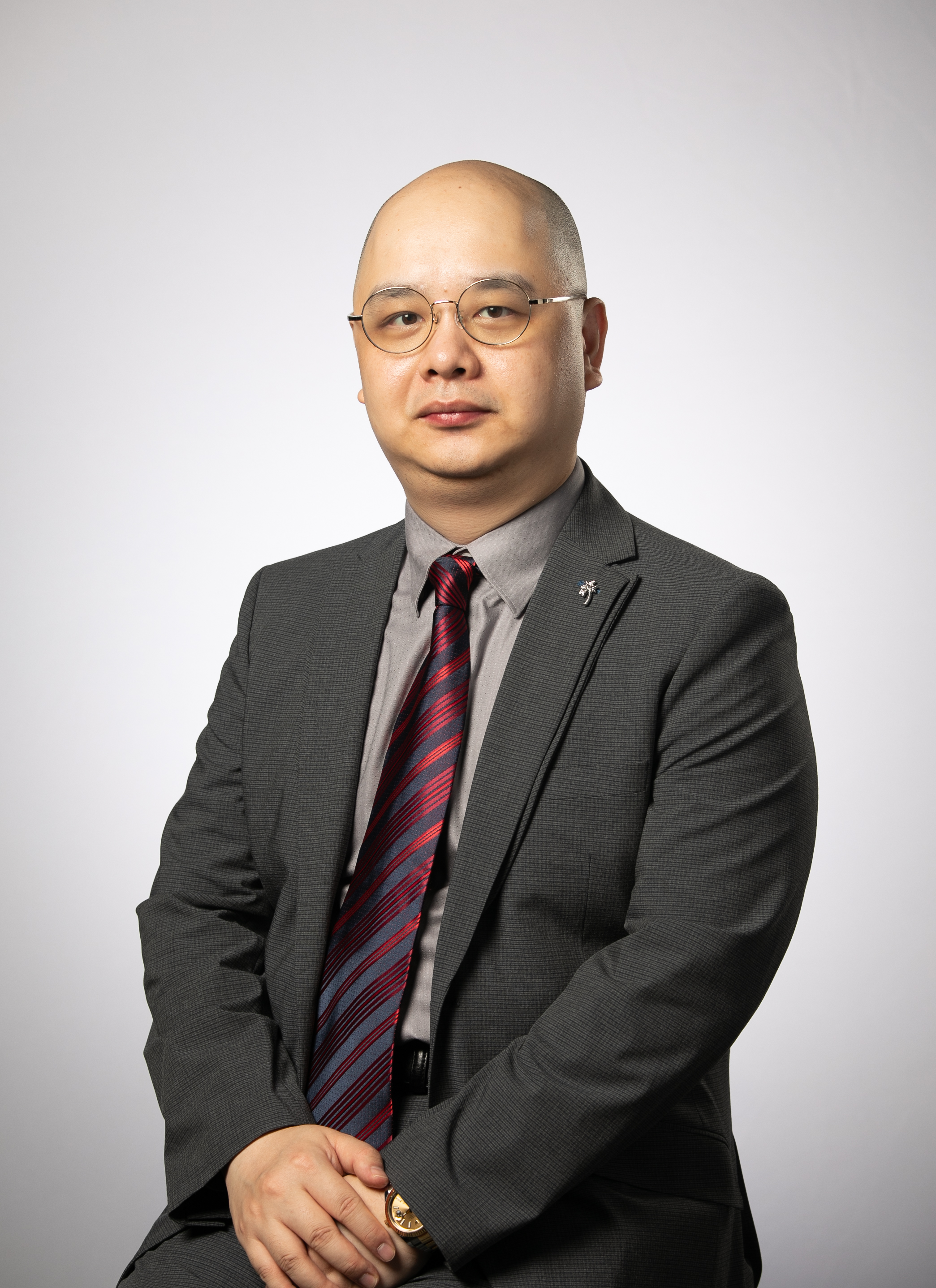}}]{Haoran Xie} (Senior Member, IEEE)
 received a Ph.D. degree in Computer Science from City University of Hong Kong and an Ed.D degree in Digital Learning from the University of Bristol. He is currently the Department Head and Associate Professor at the Department of
 Computing and Decision Sciences, Lingnan University, Hong Kong. His research interests include artificial intelligence, big data, and educational technology. He has published 393 research publications, including 224 journal articles such as IEEE TPAMI, IEEE TKDE, IEEE TAFFC, and IEEE TCVST. He is the Editor-in-Chief of Natural Language Processing Journal, Computers \& Education: Artificial Intelligence and Computers \& Education: X Reality. He has been selected listed as the World's Top 2\% Scientists by Stanford University. 
 \end{IEEEbiography}

\begin{IEEEbiography}[{\includegraphics[width=1in,height=1.25in,clip,keepaspectratio]
{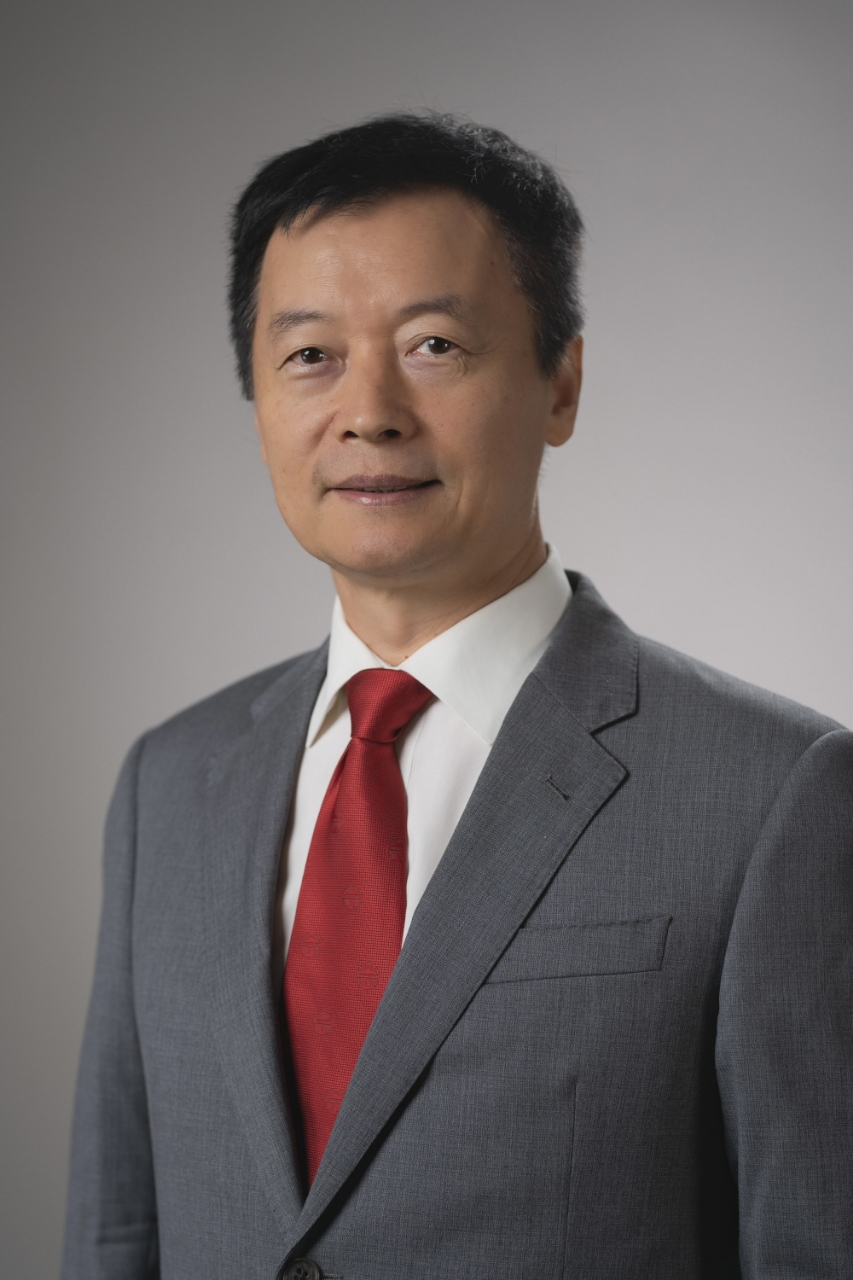}}]
{Si-Zhao Joe Qin} (Fellow, IEEE) received the B.S. and M.S. degrees in automatic control from Tsinghua University, Beijing, China, in 1984 and 1987, respectively, and the Ph.D. degree in chemical engineering from the University of Maryland, College Park, MD, USA, in 1992. He is currently the Wai Kee Kau Chair Professor and President of Lingnan University, Hong Kong. His research interests include data science and analytics, machine learning, process monitoring, model predictive control, system identification, smart manufacturing, smart cities, and predictive maintenance. 
Prof. Qin is a Fellow of the U.S. National Academy of Inventors, IFAC, and AIChE. He was the recipient of the 2022 CAST Computing Award by AIChE, 2022 IEEE CSS Transition to Practice Award, U.S. NSF CAREER Award, and NSF-China Outstanding Young Investigator Award. His h-indices for Web of Science, SCOPUS, and Google Scholar are 66, 73, and 84, respectively.
\end{IEEEbiography}

\begin{IEEEbiography}[{\includegraphics[width=1in,height=1.25in,clip,keepaspectratio]
{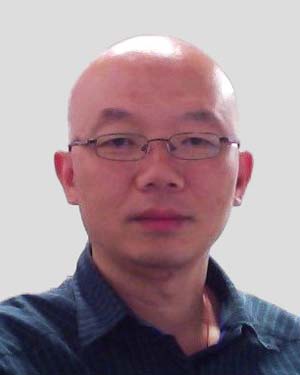}}]
{Xiaohui Tao} (Senior Member, IEEE) is currently a Full Professor with the University of Southern Queensland, Toowoomba, QLD, Australia. His research interests include artificial intelligence, data analytics, machine learning, knowledge engineering, information retrieval, and health informatics. His research outcomes have been published across more than 150 papers including many top-tier journals and conferences. He is a Senior Member of ACM and the Vice Chair of IEEE Technical Committee of Intelligent Informatics. He was the recipient of an ARC DP in 2022. He is the Editor-in-Chief of Natural Language Processing Journal.
\end{IEEEbiography}

\begin{IEEEbiography}[{\includegraphics[width=1in,height=1.25in,clip,keepaspectratio]{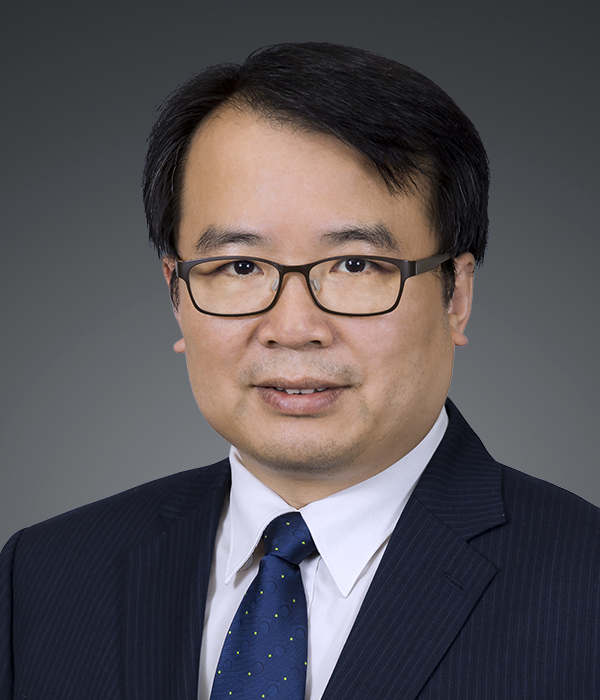}}]{Fu Lee Wang} (Senior Member, IEEE) received the B.Eng. degree in computer engineering and the M.Phil. degree in computer science and information systems from The University of Hong Kong, Hong Kong, and the Ph.D. degree in systems engineering and engineering management from The Chinese University of Hong Kong, Hong Kong. 
Prof. Wang is the Dean of the School of Science and Technology, Hong Kong Metropolitan University, Hong Kong. He has over 300 publications in international journals and conferences and led more than 20 competitive grants with a total greater than HK\$20 million. His current research interests include educational technology, information retrieval, computer graphics, and bioinformatics. 
Prof. Wang is a fellow of BCS, HKIE and IET and a Senior Member of ACM. He was the Chair of the IEEE Hong Kong Section Computer Chapter and ACM Hong Kong Chapter.
\end{IEEEbiography}

%\begin{IEEEbiography}[{\includegraphics[width=1in,height=1.25in,clip,keepaspectratio]{hndai.jpg}}]{Hong-Ning Dai} (Senior Member, IEEE)  received the Ph.D. degree in computer science and engineering from The Chinese University of Hong Kong, Hong Kong, in 2008. He is currently an Associate Professor at the Department of Computer Science, Hong Kong Baptist University, Hong Kong. Before joining Hong Kong Baptist University, he was with the School of Computer Science and Engineering, Macau University of Science and Technology, Cotai, Macau, as an Assistant Professor/Associate Professor from 2010 to 2021; and the Department of Computing and Decision Sciences, Lingnan University, Hong Kong, as an Associate Professor from 2021 to 2022. He has coauthored/co-edited four monograph books and published more than 200 peer-reviewed papers in top-tier journals and conferences. His research interests include the Internet of Things, blockchains, and big data analytics. Dr. Dai has served as an Associate Editor for IEEE Transactions on Intelligent Transportation Systems, IEEE Transactions on Industrial Informatics, and Ad Hoc Networks.
%\end{IEEEbiography}

\end{document}